\documentclass[sigconf]{acmart}

\AtBeginDocument{%
	\providecommand\BibTeX{{%
			\normalfont B\kern-0.5em{\scshape i\kern-0.25em b}\kern-0.8em\TeX}}}

\acmYear{2022}\copyrightyear{2022}
\acmConference[MobiHoc '22]{The Twenty-third International Symposium on Theory, Algorithmic Foundations, and Protocol Design for Mobile Networks and Mobile Computing}{October 17--20, 2022}{Seoul, Republic of Korea}
\acmBooktitle{The Twenty-third International Symposium on Theory, Algorithmic Foundations, and Protocol Design for Mobile Networks and Mobile Computing (MobiHoc '22), October 17--20, 2022, Seoul, Republic of Korea}
\acmPrice{15.00}
\acmDOI{10.1145/3492866.3549723}
\acmISBN{978-1-4503-9165-8/22/10}
\acmPrice{15.00}
\acmISBN{978-1-4503-XXXX-X/18/06}

\usepackage{algorithm}
\usepackage{algorithmic}
\usepackage{xcolor}
\usepackage{color}
\usepackage{inputenc} 		
\usepackage[T1]{fontenc}    
\usepackage{hyperref}       
\usepackage{url}            
\usepackage{booktabs}       
\usepackage{amsfonts}       
\usepackage{nicefrac}       
\usepackage{microtype}      
\usepackage{float}
\usepackage{algorithm}
\usepackage{makecell}
\usepackage{multirow}
\usepackage{wrapfig}
\usepackage{lipsum}
\usepackage{hhline}
\usepackage{algorithmic}
\usepackage{amsmath,bm}
\usepackage{amsthm}
\usepackage[shortlabels]{enumitem}
\usepackage{mathrsfs}
\usepackage{xcolor}
\usepackage{graphicx}
\usepackage{color}
\usepackage{soul}
\usepackage{wrapfig}
\usepackage{subfigure}
\usepackage{bbding}
\usepackage{soul}

\usepackage{pifont}
\renewcommand{\algorithmicrequire}{\textbf{Input:}}
\renewcommand{\algorithmicensure}{\textbf{Output:}}

\renewcommand{\a}{\mathbf{a}}

\newcommand{\bg}{\bar{\mathbf{g}}}

\newcommand{\D}{\mathcal{D}}

\newcommand{\Eb}{\mathbb{E}}

\newcommand{\f}{\mathbf{f}}

\newcommand{\bnf}{\overline{\nabla \f}}
\newcommand{\nf}{{\nabla \f}}

\newcommand{\Fc}{\mathcal{F}}

\newcommand{\g}{\mathbf{g}}

\newcommand{\I}{\mathbf{I}}

\newcommand{\Nc}{\mathcal{N}}

\newcommand{\Q}{\mathbf{Q}}

\newcommand{\W}{\mathbf{W}}

\newcommand{\x}{\mathbf{x}}

\newcommand{\y}{\mathbf{y}}
\newcommand{\by}{\bar{\mathbf{y}}}

\newcommand{\xb}{\mathbf{\bar{x}}}

\newcommand{\1}{\mathbf{1}}

\newtheorem{thm}{Theorem}

\newtheorem{cor}[thm]{Corollary}
\newtheorem{lem}{Lemma}

\newtheorem{rem}{Remark}

\newtheorem{assum}{Assumption}

\newcommand{\vzeta}{\boldsymbol \zeta}

\renewcommand{\algorithmicrequire}{\textbf{Input:}}

\newtheorem{innercustomgeneric}{\customgenericname}
\providecommand{\customgenericname}{}
\newcommand{\newcustomtheorem}[2]{%
  \newenvironment{#1}[1]
  {%
   \renewcommand\customgenericname{#2}%
   \renewcommand\theinnercustomgeneric{##1}%
   \innercustomgeneric
  }
  {\endinnercustomgeneric}
}

\newcustomtheorem{customthm}{Theorem}
\newcustomtheorem{customlemma}{Lemma}

\usepackage{newfloat}
\usepackage{listings}



\begin{document}

\title[NET-FLEET: Decentralized Federated Learning with Heterogeneous Data]{NET-FLEET: Achieving Linear Convergence Speedup for Fully Decentralized Federated Learning with Heterogeneous Data}

\author[Zhang et al.]{Xin Zhang$^*$, Minghong Fang$^{+}$, Zhuqing Liu$^{+}$, Haibo Yang$^{+}$, Jia Liu$^{+}$,  and Zhengyuan Zhu$^*$}
\affiliation{
	\institution{$^*$Department of Statistics, Iowa State University}
	\institution{$^+$Department of Electrical and Computer Engineering, The Ohio State University}
	\country{}
}

	%


%
%
%
%
%
%

\begin{abstract} 
Federated learning (FL) has received a surge of interest in recent years thanks to its benefits in data privacy protection, efficient communication, and parallel data processing.
Also, with appropriate algorithmic designs, one could achieve the desirable {\em linear speedup for convergence} effect in FL.
However, most existing works on FL are limited to systems with i.i.d. data and centralized parameter servers and results on decentralized FL with heterogeneous datasets remains limited.
Moreover, whether or not the linear speedup for convergence is achievable under {\em fully decentralized} FL with data heterogeneity remains an open question.
In this paper, we address these challenges by proposing a new algorithm, called NET-FLEET, for fully decentralized FL systems with data heterogeneity.
The key idea of our algorithm is to enhance the local update scheme in FL (originally intended for communication efficiency) by incorporating a recursive gradient correction technique to handle heterogeneous datasets.
We show that, under appropriate parameter settings, the proposed NET-FLEET algorithm achieves a linear speedup for convergence.
We further conduct extensive numerical experiments to evaluate the performance of the proposed NET-FLEET algorithm and verify our theoretical findings.
\end{abstract}

\maketitle

\section{Introduction}\label{Section: introduction}

Federated learning (FL) is a powerful distributed training paradigm for modern large-scale machine learning  \cite{yang2019federated,li2020federated,kairouz2019advances,xu2020federated,lu2020decentralized,brisimi2018federated,cao2020fltrust,kang2020reliable,yang2021cfedavg,khanduri2021achieving,yang2022anarchic}.
FL leverages a large number of workers to collaboratively learn a global model.
Mathematically, FL aims to solve an optimization problem in the form of:
\begin{align}\label{Eq: FL problem}
\min_{\x \in \mathbb{R}^{p}} f(\x) \triangleq \frac{1}{m}\sum_{i=1}^{m} f_i(\x),
\end{align}
where $f_i(\x) \triangleq \Eb_{\vzeta\sim \mathcal{D}_{i}}[f_i(\x;\vzeta_i)]$ is the loss function of the data distribution $\mathcal{D}_{i}$ at worker $i$, and $m$ is the number of workers. 
Different from traditional learning algorithms where data are collected and stored in a centralized server, FL allows the training data distributed at the workers, which could be smart phones, robots, network sensors, or other local information sources.
A global model can be trained without the need to share the workers' data over the network, thus helping preserve data privacy.
However, FL also faces several major technical challenges:

\begin{list}{\labelitemi}{\leftmargin=0.2em \itemindent=2.5em \itemsep=.0em}
\item[(C1).] {\textbf{Data Heterogeneity:}} 
In conventional distributed learning, the data are either globally available or randomly shuffled and assigned to each worker.
Thus, it is safe to assume that the data distributions at the workers are identical, i.e. $\mathcal{D}_i =\mathcal{D}_j$, $\forall i\in[m]$.
Unfortunately, in FL systems, data are generated locally at each worker based on their own circumstances.
As a result, data heterogeneity among the workers is unavoidable.
Such data heterogeneity imposes significant challenges in designing FL algorithms and their training performance analysis.

\item[(C2).] {\textbf{Unreliable Centralized Server:}} 
Most current distributed learning systems are based on the server-worker architecture, where workers are coordinated by a centralized server. 
However, the centralized server may suffer several limitations, e.g., vulnerability to cyber-attacks and being a significant communication bottleneck. 
Additionally, in the context of FL, it is sometimes hard or even infeasible to find a trustworthy centralized server with whom all workers are willing to share information.
\end{list}

The above key challenges motivate us to consider {\em fully decentralized} FL systems (i.e., {\em without} any centralized server) deployed over peer-to-peer networks.
Toward this end, in this paper, we focus on the fundamental ``linear speedup for convergence'' problem for decentralized FL under data heterogeneity.
In the literature, it is well-known that the centralized-server-aided FL enjoys the  ``linear speedup for convergence'' property.
Specifically, the work in \cite{stich2018local,yu2019parallel} showed that the celebrated FedAvg algorithm and its variants under the homogeneous data setting can achieve a convergence rate of $O(1/\sqrt{mKS})$ with a sufficiently large communication rounds $S$, where $m$ is the number of workers and $K$ is the number of local update rounds.
Notably, the $O(1/\sqrt{mKS})$ convergence rate implies a ``linear speedup'' with respect to the number of workers $m$.
This is because, to attain an $\epsilon$-accuracy in convergence, an algorithm with a convergence rate $\mathcal{O}(1/\sqrt{S})$ takes $\mathcal{O}(1/\epsilon^2)$ steps.
In contrast, an algorithm with a convergence rate $\mathcal{O}(1/\sqrt{mS})$ needs $\mathcal{O}(1/m \epsilon^2)$ steps (the hidden constant in Big-O is the same). 
In this sense, the convergence rate $\mathcal{O}(1/\sqrt{mS})$ implies a {\em linear speedup} with respect to the number of workers.
Such a linear speedup is highly desirable because it implies that one can efficiently leverage the massive parallelism in large-scale FL systems.
However, under the data heterogeneity and unreliable centralized server challenges outline in (C1-C2), a fundamental open question arises: 
{\em Can we still achieve the state-of-the art linear speedup for convergence, i.e., $O(1/\sqrt{mKS})$, under a fully decentralized FL system with data heterogeneity?}

In this paper, we give an {\em affirmative} answer to this question and propose a new {\em recursive gradient correction} based fully decentralized FL algorithm.
Our main contributions are summarized as follows:

\begin{list}{\labelitemi}{\leftmargin=1em \itemindent=0em \itemsep=.2em}

\item 
To circumvent the unreliable centralized server challenge, we propose a fully decentralized network FL algorithm called {\em  Decentralized \underline{Net}worked \ul{F}ederated \ul{Le}arning with Recursive Gradi\ul{e}nt Correc\ul{t}ion (NET-FLEET)}.
In NET-FLEET, there is no centralized server and workers only need to share information with their neighboring nodes in each communication round.
Similar to FedAvg-type algorithms, our proposed NET-FLEET algorithm allows the workers to run multiple local updates between two consecutive communication rounds with their neighbors, so as to reduce the communication load. 
By eliminating the centralized server, our NET-FLEET algorithm achieves gains in both robustness and flexibility.

\item 
By proposing a new {\em recursively corrected stochastic gradient estimator} technique, our NET-FLEET algorithm works with decentralized network systems where workers hold heterogeneous datasets.
It is worth noting that, although the conventional gradient tracking method \cite{pu2020distributed,xin2020improved,qu2017harnessing} shares some similarity with our  technique, the conventional gradient tracking method {\em cannot} be directly adopted in decentralized FL since the gradient estimators for local updates are not clearly defined in conventional gradient tracking. 
In contrast, our new corrected gradient estimator efficiently approximates the global stochastic gradient, so that it can handle data heterogeneity in decentralized FL.

\item 
We establish theoretical guarantees for the convergence performance of NET-FLEET. 
The key challenge in the analysis is to examine the local model consensus error caused by {\em multiple} local updates contained in one round of fully decentralized model averaging.
So far, most theoretical results in the FL literature rely on the assumption of homogenous datasets or gradient dissimilarity conditions.
In this work, we relax these conditions and show that our proposed algorithm enjoys an $O(1/\sqrt{mSK})$ convergence rate with {\em arbitrary} heterogeneous datasets.
Our result implies a {\em linear speedup} for convergence with respect to the worker number.
Notably, our analysis and convergence results do not require the bounded gradient and homogeneous data assumptions, which could be of independent interest to general non-convex FL problems. 


\end{list}
Collectively, our results in this paper contribute to the state of the art of decentralized FL with data heterogeneity.
 The rest of the paper is organized as follows.
 In Section~\ref{sec:related}, we review the literature to put our work in comparative perspectives.
 In Section~\ref{sec:algorithm}, we formally state decentralized FL problem and propose our {NET-FLEET} algorithm.
 The convergence rate and complexity analysis of our algorithms are provided in Section~\ref{sec:analysis}.
 We provide numerical results in Section~\ref{sec:numerical} to verify the theoretical results of our algorithms.
 In Section~\ref{sec:conclusion}, we provide concluding remarks and discussions.

\section{Related Work}\label{sec:related}

In this section, we provide a quick overview on recent related work on FL algorithms with homogeneous and heterogeneous datasets, as well as algorithms for fully decentralized FL in the literature.

{\bf 1) FL with Homogenous Datasets:} 
The federated averaging (FedAvg) algorithm, also known as ``Local SGD,'' was first developed by \cite{mcmahan2017communication} as a heuristic approach to address FL.
FedAvg lets workers run $K$ successive SGD updates with local data before communicating with the central server, thus achieving better communication efficiency than the traditional parallel SGD.
Since then, FedAvg has sparked a large number of follow-ups that focus on theoretical performance of FL with homogeneous data (see, e.g., \cite{stich2018local,yu2019parallel,wang2018cooperative,stich2020error,lin2018don}).
Under the homogeneous data assumption, most of the works provide a linear speedup for convergence, i.e. an $O(1/\sqrt{mSK})$, for a sufficiently large communication rounds $S$, which matches the state-of-the-art convergence rate of the parallel SGD \cite{dekel2012optimal,ghadimi2013stochastic}.
Furthermore, it has also been shown in \cite{lin2018don} that FedAvg enjoys a better generalization performance than parallel SGD.
We refer readers to excellent recent surveys \cite{li2020federated,kairouz2019advances} for a comprehensive review.

{\bf 2) FL with Heterogeneous Datasets:}
More recently, researchers have started to investigate the performance of FedAvg and its variants for FL with heterogeneous datasets. 
The work in \cite{zhao2018federated} first showed that the accuracy of FL degrades significantly for neural networks trained on highly skewed heterogeneous datasets. 
They explained such accuracy degradation by the weight divergence, which can be quantified by the Wasserstein distance between the population data distributions and the workers' data distributions.
To mitigate such worker-drift effects, they proposed a strategy to improve training with heterogeneous data by sharing a small subset of data between all the workers.
So far, most of the existing theoretical work in the literature (see, e.g., \cite{yu2019parallel,wang2019adaptive,sahu2018convergence,haddadpour2019convergence}) analyzed FedAvg's worker-drift with a $(G,B)$-bounded gradient dissimilarity assumption (GBD assumption), i.e., $\frac{1}{m}\sum_{i=1}^{m} \|\nabla f_i(\x) - \nabla f(\x)\|^2 \le G^2 + B^2\|\nabla f(\x)\|^2$, $\forall i\in[m]$.
With the $(G,B)$-GBD assumption.
These works showed that FedAvg could achieve a linear speedup for convergence with the rounds of local updates $K$ being $\sqrt[3]{S}/m$.
To relax the extra assumption on gradients, the work in \cite{liang2019variance} proposed a Variance Reduced Local SGD (VRL-SGD) algorithm for FL with heterogeneous data.
VRL-SGD introduces an auxiliary variable to track average deviation between the local gradients and the corresponding global gradient of the same model parameters, and uses it to approximate the global gradients during the local SGD updates.

To further reduce the communication complexity, the work in \cite{yang2021achieving} recently developed a generalized FedAvg (G-FedAvg) algorithm with two-sided learning rates and improved $K$ to be as large as $S/m$.
In G-FedAvg,  the workers first run local updates with a local step-size, then upload the local parameter changes to the centralized server. 
Upon receiving workers' information, the server updates the global model parameter with the local changes and a server-side step-size.
Due to the two-sided learning rates, the G-FedAvg achieved a linear speedup for convergence with a large $K$.
But their analysis and convergence results are still limited by the dissimilarity of local gradients.
The work in \cite{karimireddy2020scaffold} proposed a Stochastic Controlled Averaging (SCAFFOLD) algorithm, which corrects the worker-drift problem also by utilizing two-sided learning rates and control variables.
SCAFFOLD estimates the worker-drift by the difference between the server-side  and worker-side control variables and uses it to correct the local update.
After $K$ rounds of local updates, the workers send the local parameter changes to the centralized server for server-side update.
By using the two-sided step-sizes and control variables, SCAFFOLD achieves a linear speedup for convergence without making assumptions on gradients.
However, the aforementioned algorithms only work for the systems with a centralized parameter server. 

{\bf 3) Decentralized FL Algorithms:}
Decentralized FL has also received increasing attention recently, which is motivated by the fact that in some FL scenarios, the centralized server is not trustable.
For example, the work in \cite{li2019communication} proposed a Local Decentralized SGD (LD-SGD) algorithm for decentralized FL.
LD-SGD can be viewed as a variant of the well-known Decentralized SGD (DSGD) algorithm \cite{nedic2009distributed,yuan2016convergence,zeng2018nonconvex,lian2017can}.
In LD-SGD, the workers perform multiple local updates and then communicate with their neighbors to perform one round of parameter aggregation.
It is shown that LD-SGD could achieve a linear speedup for convergence under the bounded gradient assumption. 
Recently, the work in \cite{gao2020periodic} developed a periodic decentralized momentum SGD (PD-SGDM) algorithm, which uses the gradient momentum term to improve the convergence performance.
With a bounded gradient assumption, PD-SGDM can achieve a linear speedup for convergence as long as the rounds of local updates is bounded by $K={\sqrt[3]{S}}/{m}$, which matches the number of local updates of the FedAvg algorithm. 
The work in \cite{yu2019linear} also proposed a decentralized momentum SGD algorithm with local updates.
Unlike the PD-SGDM which assumes the bounded gradient, \cite{yu2019linear} leverages the generalized GBD assumption to handle the data heterogeneity and achieve the same linear speedup.
In this work, we aim to achieve a linear speedup for decentralized federated learning without any assumption on gradient boundedness.

The most related work to our NET-FLEET is the decentralized FL stochastic gradient tracking (DSGT) algorithm proposed by \cite{lu2020decentralized}.
In DSGT, the workers first run $K$ rounds local SGD updates and then perform one round of stochastic gradient tracking update.
However, the authors only provided a convergence analysis for the case with $K=0$, i.e., no local update. 
In comparison, our NET-FLEET algorithm employs a local update scheme with a new {\em recursive gradient correction} technique.
We show that NET-FLEET achieves a linear speedup for convergence with local updates rounds $K = \sqrt[3]{S}/m$ {\em without} any bounded gradient assumption.



\section{Problem Statement and Algorithm Design}\label{sec:algorithm}

In this section, we will first state the fully decentralized FL problem.
Then, we will present our NET-FLEET algorithm.

\subsection{Decentralized Federated Learning}

In the fully decentralized FL scenario, the workers form a peer-to-peer network system, which can be represented by an undirected connected graph $\mathcal{G} = (\mathcal{N}, \mathcal{L})$.
Here, $\mathcal{N}$ and $\mathcal{L}$ are the sets of workers and edges, respectively, with $|\mathcal{N}| = m$. 
The workers are capable of local computation and communicating with their neighboring workers via the edges in $\mathcal{L}$.
The goal of fully decentralized FL is to have the workers {\em distributively} and {\em collaboratively} solving the global optimization problem in the following form:
\begin{align}\label{Eq: general_problem}
\min_{\x \in \mathbb{R}^p} f(\x) = \min_{\x \in \mathbb{R}^p} \frac{1}{m}\sum_{i=1}^{m} f_i(\x),
\end{align}
where each local objective function $f_i(\x) \triangleq \Eb_{\zeta\sim\D_i} f_i(\x;\zeta)$ is only observable to worker $i$ and not necessarily convex.
Here, $\D_i$ represents the distribution of the dataset at node $i$, which is {\em heterogeneous} across workers.
To solve Problem~\eqref{Eq: general_problem} in a decentralized fashion, one can reformulate Problem~(\ref{Eq: general_problem}) in the following equivalent form by introducing a local model copy at each worker:
\begin{align}\label{Eq: consensus_problem}
& \text{Minimize} && \hspace{-.5in} \frac{1}{m}\sum_{i=1}^{m} f_i(\x_i) & \\
& \text{subject to} && \hspace{-.5in} \x_i = \x_j, && \hspace{-.5in} \forall (i,j) \in \mathcal{L}. \nonumber
\vspace{-.05in}
\end{align}  
where $\x \triangleq [\x_1^\top,\cdots,\x_m^\top]^\top,$ and $\x_i$ is an introduced local copy at worker $i$.
To solve Problem \eqref{Eq: consensus_problem}, we consider an $\epsilon^2$-stationary point $\x$ defined as follows:
\begin{align}\label{Eq: FOSP_network}
\underbrace{\Big\|\frac{1}{m}\sum_{i=1}^{m} \nabla f_i(\xb) \Big\|^2}_{\mathrm{Global \,\, gradient \,\, magnitude}} \!\!\!\! + \underbrace{\frac{1}{m}\sum_{i=1}^{m}\|\x_{i}- \xb\|^2}_{\mathrm{Consensus \,\, error}} \le \epsilon^2,
\end{align}
where $\xb \triangleq \frac{1}{m}\sum_{i=1}^{m} \x_{i}$ represents the global average across all workers.
Unlike the $\epsilon^2$-stationary point for centralized FL, the above criterion in Eq.~\eqref{Eq: FOSP_network} includes two components: the first term is the gradient norm of the global loss function and the second term is the average consensus error across all local copies.
In this work, we aim to develop an efficient algorithm to attain an $\epsilon^2$-stationary point for fully decentralized FL with heterogeneous datasets and study its speedup performance as the number of workers increases. 

\subsection{The NET-FLEET Algorithm}

Now, we present our {Decentralized \ul{Net}worked \ul{F}ederated \ul{Le}arning with Recursive Gradi\ul{e}nt Correc\ul{t}ion (NET-FLEET)} algorithm. 
To solve Problem~\eqref{Eq: FL problem} in decentralized network systems where workers reach a {\em consensus} on a global optimal solution, a common approach in the literature is to let workers aggregate neighboring information through a consensus matrix $\W \in \mathbb{R}^{m\times m}$.
Let $[\W]_{ij}$ represent the element in the $i$-th row and the $j$-th column in $\W$.
Then, a consensus matrix $\W$ should satisfy the following properties:
\begin{enumerate}[topsep=1pt, itemsep=-.1ex, leftmargin=.35in]
\item[(a)] {\em Doubly Stochastic:} $\sum_{i=1}^{m} [\mathbf{W}]_{ij}=\sum_{j=1}^{m} [\mathbf{W}]_{ij}=1$.
\item[(b)] {\em Symmetric:} $[\mathbf{W}]_{ij} = [\W]_{ji}$, $\forall i,j \in \mathcal{N}$. 
\item[(c)] {\em Network-Defined Sparsity Pattern:} $[\W]_{ij} > 0$ if $(i,j)\in \mathcal{L};$ otherwise $[\mathbf{W}]_{ij}=0$, $\forall i,j \in \mathcal{N}$.
\end{enumerate}
The above properties imply that the eigenvalues of $\W$ are real and can be sorted as $-1 < \lambda_m(\W) \leq \cdots \leq \lambda_2(\W) < \lambda_1(\W) = 1$.
We define the second-largest eigenvalue in magnitude of $\W$ as $\lambda \triangleq \max\{|\lambda_2(\W)|,|\lambda_m(\W)|\}$ for further notation convenience.
It can be seen later that $\lambda$ plays an important role in the step-size selection and characterizing the algorithm's convergence rate.

Similar to the centralized-server-based FL, a key defining feature in decentralized FL is that it allows workers to update the local model parameters {\em multiple rounds} before workers' communication and model averaging.
However, with heterogeneous data at different workers, the update directions (i.e., the stochastic gradients) are not identically distributed.
Thus, after several local update rounds, the local parameters will move towards their worker-side optimum $\x^{*(i)}$, where $\x^{*(i)} = \arg \min f_i(\x)$.
This phenomenon may cause divergence of the algorithm and is often referred to as the ``worker-drift problem.'' 
Moreover, the lack of a centralized sever further worsens the worker-drift problem. 
To address this challenge, in our NET-FLEET algorithm, we introduce an auxiliary parameter $\y^{(i)}$ at each worker $i$ to approximate the global stochastic gradients.
Our NET-FLEET algorithm is illustrated in Algorithm~\ref{Algorithm: NET-FLEET}.

Specifically, NET-FLEET has $K$ inner loops at each worker for the local updates between two consecutive outer loop iterations for inter-worker communications.
Also, there are $S$ rounds of inter-worker communications.
At each outer loop iteration $s$, workers share the local model parameter $\x_{s,0}^{(i)}$ and the corrected gradient parameter $\y_{s,0}^{(i)}$ with neighboring workers, and initialize the inner-loop's starting points as $\x_{s,1}^{(i)}$ and $\y_{s,1}^{(i)}$ based on the neighboring average and local stochastic gradient update.
Then, within the local inner loops, the update of $\y^{(i)}$ follows a recursive structure:
\begin{align}\label{Eq:gradient tracking}
\y_{s,k+1}^{(i)}=   \y_{s,k}^{(i)}  + \g_{s,k+1}^{(i)}- \g_{s,k}^{(i)},~~ \forall k \in 1, \cdots, K-1,
\end{align}
where $s$ and $k$ are the indices of outer and inner loops, respectively, and $ \g_{s,k}^{(i)} = \nabla f_i(\x_{s,k}^{(i)};\vzeta_{s,k}^{(i)})$ is the local stochastic gradient with random sample $\vzeta_{s,k}^{(i)}$.
In~\eqref{Eq:gradient tracking}, it can be easily verified that the correction term follows $\y_{s,k}^{(i)}  - \g_{s,k}^{(i)} = \y_{s,1}^{(i)} - \g_{s,1}^{(i)} = {{\sum_{j\in \Nc_i} [\W]_{ij} \y_{s,0}^{(i)}}} - \g_{s,0}^{(i)}$, which measures the difference between the local stochastic gradient and neighboring weighted-average update direction.
By adding such correction term to $\g_{s,k+1}^{(i)}$, $\y_{s,k+1}^{(i)}$ will be close to the global stochastic gradient as outer loop iteration $s$ gets large.
Note that in NET-FLEET, the model parameter $\x$ is updated $SK$ times, but the number of information communication rounds between workers is only $S$ times. 
Thus, compared with traditional decentralized learning algorithms, NET-FLEET reduces the overall communication cost by a $1/K$ factor.

\begin{rem}
{\em 
Some important remarks regarding our recursive gradient correction technique are in order.
First, we note that the idea of gradient correction has appeared in the literature, including stochastic variance reduction (SVR) method in SVRG\cite{johnson2013accelerating}/SPIDER\cite{fang2018spider}, gradient tracking (GT) method in GNSD\cite{lu2019gnsd}/GT-DSGD\cite{xin2020improved}, etc.
However, the key differences between our method and these existing works are: 
1) The SVR method requires a precise global gradient estimation at each outer loop iteration, while in our method the outer loops' gradient estimator is based on an {\em inexact} neighboring averaging and recursive correction;
2) The GT method is designed with a single-loop structure and demands one round of communication after each local update, thus suffering {\em high communication costs}. 
This limitation is due to the iterates' contraction result in the conventional convergence analysis for the GT method (cf. \cite[Lemma~3]{lu2019gnsd}), which does not hold for multiple local updates.
In contrast, our new recursive gradient correction method works with {\em multiple} local updates under decentralized FL.
In this sense, the GT method is a special case of our method when local updates $K = 1$.
}
\end{rem}

      \begin{algorithm}[t!]
        \caption{The NET-FLEET Algorithm.}\label{Algorithm: NET-FLEET}.
        \begin{algorithmic}[1]
        \REQUIRE Initial point $\x^0$, learning rate $\eta$, communication rounds $S$, local update rounds $K$.
        \STATE Set $\x_{0,0}^{(i)} = \x^0$ and $\y_{0,0}^{(i)} = \g_{0,0}^{(i)} =  \nabla f_i(\x_{0,0}^{(i)};\vzeta_{0,0}^{(i)})$ at worker $i$, for all $i\in[m]$.
\FOR{$s = 0, \cdots, S-1$}
\FOR{worker $i$, $i \in [m]$}
	\STATE Share $(\x_{s,0}^{(i)}, \y_{s,0}^{(i)})$ with neighboring nodes;
    \STATE Update $\x_{s,1}^{(i)}  = {{\sum_{j\in \Nc_i} [\W]_{ij} \x_{s,0}^{(j)}}} - \eta \y_{s,0}^{(i)}$;
    \STATE Calculate $\g_{s,1}^{(i)} =  \nabla f_i(\x_{s,1}^{(i)};\vzeta_{s,1}^{(i)})$;
	\STATE Correct $\y_{s,1}^{(i)} = {{\sum_{j\in \Nc_i} [\W]_{ij} \y_{s,0}^{(j)}}}  +\g_{s,1}^{(i)} - \g_{s,0}^{(i)}$;
	\FOR{$k = 1, \cdots, K-1$}
    	\STATE Update $\x_{s,k+1}^{(i)} =  \x_{s,k}^{(i)} - \eta \y_{s,k}^{(i)}$;
    	\STATE Calculate $\g_{s,k+1}^{(i)} =  \nabla f_i(\x_{s,k+1}^{(i)};\vzeta_{s,k+1}^{(i)})$;
		\STATE Correct $\y_{s,k+1}^{(i)}=   \y_{s,k}^{(i)}  + \g_{s,k+1}^{(i)}- \g_{s,k}^{(i)}$
	\ENDFOR
	\STATE  Set $\x_{s+1,0}^{(i)} \!=\! \x_{s,K}^{(i)}$, $\y_{s+1,0}^{(i)}\! =\! \y_{s,K}^{(i)}$, $\g_{s+1,0}^{(i)} \!=\! \g_{s,K}^{(i)}$;
\ENDFOR
\ENDFOR
        \end{algorithmic}
      \end{algorithm}

\section{Theoretical Performance Analysis}\label{sec:analysis}

In this section, we will establish the convergence properties of our proposed NET-FLEET algorithm.
Due to space limitation, we outline the key steps of the proofs of Theorem \ref{thm:Convergence of NET-FLEET}. 
We relegate the proof details to the supplementary material.
We start with stating the following assumptions:
\begin{assum}\label{Assumption: function}
The objectives $f(\cdot)$ and $f_i(\cdot)$ satisfy:
\begin{enumerate}[topsep=1pt, itemsep=-.1ex, leftmargin=.2in]
	\item[(1)] $f(\x)$ is bounded from below, i.e., there exists an $\x^*\in\mathbb{R}^p,$ such that $f(\x)\ge f(\x^*)$, $\forall \x\in\mathbb{R}^p;$
	\item[(2)] The function $f_i(\x)$ is continuously differentiable and has $L$-Lipschitz continuous gradients, i.e., there exists a constant $L >0$ such that $|\nabla f_i(\x_1) -\nabla f_i(\x_2)|\le L \| \x_1-\x_2 \|_2,$ $\forall \x_1,\x_2\in\mathbb{R}^p;$
	\item[(3)] The stochastic gradient is unbiased and has bounded variance with respect to the local data distribution, i.e., $\Eb_{\vzeta \sim\D_i}[\nabla f_i(\x;\vzeta)] = \nabla f_i(\x)$ and $\text{Var}_{\vzeta \sim \D_i}[\nabla f_i(\x;\vzeta)]\le \sigma^2$ for some constant $\sigma >0$. 
\end{enumerate}
\end{assum}
It is worth noting that we do not need the conventional bounded gradient variability assumption in most of the literature of FL with non-i.i.d. datasets.
To analyze the algorithm convergence, we define a potential function as 
\begin{align}
\mathfrak{P}_{s,k} \!\triangleq\! f(\xb_{s,k}) \!+ \!\frac{1}{m^2K}\!\sum_{i=1}^{m}(\|\x_{s,k}^{(i)}\!-\!\xb_{s,k}\|^2\!+\! C_1\eta^2\|\y_{s,k}^{(i)}\!-\!\by_{s,k}\|^2), \notag
\end{align}
where $C_1 = 6(1+\lambda K -\lambda)K/(1-\lambda)^2$ and $\xb_{s,k} = \frac{1}{m}\sum_{i=1}^{m}\x_{s,k}^{(i)}$, $\by_{s,k} = \frac{1}{m}\sum_{i=1}^{m}\y_{s,k}^{(i)}$.
With the above assumptions and definitions, we are now in a position to present the main convergence result for our NET-FLEET algorithm as follows:
\begin{thm}[Convergence of NET-FLEET]\label{thm:Convergence of NET-FLEET}
Under Assumption~\ref{Assumption: function}, if the step-size $\eta$ in Algorithm~\ref{Algorithm: NET-FLEET} satisfies:
\begin{align*}
&\eta \le \min\Big\{\frac{1}{3L}, \frac{1}{mL^2K^2}, 
\frac{(1 - \lambda)}{\sqrt{12(1 + \lambda K - \lambda)KL^2}}, \sqrt{\frac{1\!-\!\lambda}{24L^2K^2}}, \nonumber\\
&\frac{(1\!-\!\lambda)^3mK}{144}, 
\sqrt{\frac{m(1\!-\!\lambda)^2}{144LK^2}}, \frac{(1\!-\!\lambda)}{3(1\!+\!\lambda K \!-\!\lambda)mKL^2}, \frac{(1\!-\!\lambda)^2(1\!+\!\lambda K \!-\!\lambda)m}{144K}
\Big\},
\end{align*}
then Algorithm~\ref{Algorithm: NET-FLEET} has the following convergence result:
\begin{multline}\label{Eq: convergence error NET-FLEET}
\frac{1}{SK} \!\sum_{s\!=\! 0}^{S\!-\!1}\sum_{k\!=\! 0}^{K\!-\!1}\Eb\big[\|\nabla f(\xb_{s,k})\|^2 \!+\!\frac{L^2}{m}\sum_{i=1}^{m}\|\x_{s,k}^{(i)}-\xb_{s,k}\|^2\big] \leq \\
\frac{2\Eb[\mathfrak{P}_{0,0}\!-\!\mathfrak{P}_{S,0}]}{SK\eta} \!+\! \frac{3L\sigma^2\eta}{m}  
 \!+\! \frac{72\eta\sigma^2}{(1\!-\!\lambda)^2m}\!+\! \frac{72(1\!+\!\lambda K \!-\!\lambda )\eta\sigma^2}{(1\!-\!\lambda)^3Km}.
\end{multline}
\end{thm}
Several important remarks for Theorem~\ref{thm:Convergence of NET-FLEET} are in order. 
First, the convergence metric in Theorem~\ref{thm:Convergence of NET-FLEET} is $\|\nabla f(\xb_{s,k})\|^2 \!+\!\frac{L^2}{m}\sum_{i=1}^{m}\|\x_{s,k}^{(i)}-\xb_{s,k}\|^2$, where the first term is the global gradient magnitude for the non-convex objectives and the second term is the average consensus error across all local parameters in the network system.
Although depending on the Lipschitz constant $L$, this metric does not lose generality because we can change the metric to be problem instance-independent by removing $L^2$ from the second term, which is due to $\|\nabla f(\xb_{s,k})\|^2 \!+\!\frac{1}{m}\sum_{i=1}^{m}\|\x_{s,k}^{(i)}-\xb_{s,k}\|^2 \le \frac{1}{\min\{1,L^2\}}(\|\nabla f(\xb_{s,k})\|^2 \!+\!\frac{L^2}{m}\sum_{i=1}^{m}\|\x_{s,k}^{(i)}-\xb_{s,k}\|^2)$. 
With the metric in Theorem~\ref{thm:Convergence of NET-FLEET} going to zero, we have that all local parameters will asymptotically be equal and reach a first-order stationary point of the  global objective function $f(\cdot)$.
Moreover, Theorem~\ref{thm:Convergence of NET-FLEET} provides a finite-time convergence rate guarantee for our NET-FLEET algorithm.

Second, for the convergence error on the right-hand-side (RHS) of Eq.~\eqref{Eq: convergence error NET-FLEET}, with simple derivations, the first term can be bounded as: 
\begin{align*}
\frac{2}{SK\eta}\Eb[\mathfrak{P}_{0,0}\!-\!\mathfrak{P}_{S,0}]  \!\le\! 
 \frac{2}{SK\eta} \left[f(\x^0) \!+\! \frac{C_1\eta^2}{m^2K}
\!\!\sum_{i\!=\!1}^{m}\|\y_{0,0}^{(i)} \!-\! \by_{0,0}\|^2\!-\! f(\x^*)\right],
\end{align*}
which is dependent on the initialization.
The third and fourth terms are affected by the network topology: a sparser network (i.e., $\lambda$ is closer to $1$) will have larger values in these two terms.

Third, the range of step-size $\eta$ is also dependent on the network topology. A sparse network leads to a smaller step-size. 
In the following, we show that by properly selecting the parameters, our proposed NET-FLEET can achieve a linear speedup for convergence:
\begin{cor}[Linear Speedup]\label{cor: linear speedup}
Under Assumption~\ref{Assumption: function}, by setting $K = S^{1/3}/m$ and $\eta = O(\sqrt{m/SK})$, if the numbers of global and local communication rounds are sufficiently large such that $SK\ge m^{1/3}$, then NET-FLEET has the following convergence rate:
\begin{multline}\label{Eq: Linearup}
\frac{1}{SK} \sum_{s\!=\! 0}^{S\!-\!1}\sum_{k\!=\! 0}^{K\!-\!1}\Eb\big[\|\nabla f(\xb_{s,k})\|^2 \!+\!\frac{L^2}{m}\sum_{i=1}^{m}\|\x_{s,k}^{(i)}-\xb_{s,k}\|^2\big] \\
\stackrel{}{=}  O\Big(\frac{\Eb[\mathfrak{P}_{0,0}\!-\!\mathfrak{P}_{S,0}]}{\sqrt{SKm}} + \frac{\sigma^2}{\sqrt{SKm}} \Big),
\end{multline}
which implies a linear speed up for convergence. 
\end{cor}
It is worth noting that our algorithm achieves the same $K=S^{1/3}/m$ number of local updates as in \cite{gao2020periodic} {\em without} any bounded gradient assumption.

\subsection{Proof Sketch of Theorem \ref{thm:Convergence of NET-FLEET}}\label{sec:proof sketches}

Due to space limitation, we provide a proof sketch of Theorem \ref{thm:Convergence of NET-FLEET} and relegate the proof details to our online technical report\cite{proof}.
For better readability, in this section, we organize the proof of Theorem~\ref{thm:Convergence of NET-FLEET} into several key lemmas. 
Our first step to prove Theorem~\ref{thm:Convergence of NET-FLEET} is to show the descent property of our NET-FLEET algorithm, which is stated in the following lemma:

\begin{lem}\label{Lem: Descend}
Under Assumption~\ref{Assumption: function}, the following inequality holds for any outloop $s$ in Algorithm~\ref{Algorithm: NET-FLEET}:
\begin{align}\label{eqn:lemma2}
&\Eb[f(\xb_{s,K}) \!-\!f(\xb_{s,0})]
\notag\\&\!\le\! 
-\! \frac{\eta}{2} \sum_{k\!=\! 0}^{K-1}\Eb[\|\nabla f(\xb_{s,k})\|^2] 
 \!-\! \frac{\eta}{2} \sum_{k= 0}^{K-1}\Eb\bigg[\|\frac{1}{m}\sum_{i=1}^{m}\nabla f_i(\x^{(i)}_{s,k})\|^2\bigg] 
 \!\notag\\
 &+\! \frac{L\eta^2}{2} \sum_{k= 0}^{K-1}\Eb[\|\bg_{s,k}\|^2] \!+\! \frac{L^2\eta}{2m} \sum_{k= 0}^{K-1}\sum_{i=1}^{m}\Eb \bigg[\|\x^{(i)}_{s,k}\!-\!\xb_{s,k}\|^2 \bigg].
\end{align}
\end{lem}
Although Lemma~\ref{Lem: Descend} appears to be similar to conventional analysis, its proof is highly non-trivial.
In~\eqref{eqn:lemma2}, we focus on the descending upper bound for each two outloop local model parameters, between which have $K$ inner loop SGD updates, while the conventional analysis on gradient tracking method studies on two successive local model parameters with only one round of SGD update.
More Specifically, we note that the RHS of \eqref{eqn:lemma2} contains the consensus error of local model parameters $\sum_{k= 0}^{K-1}\sum_{i=1}^{m}\Eb[\|\x^{(i)}_{s,k}-\xb_{s,k}\|^2 ]$, which sums across not only the worker number $m$ but also inner loop iterations $K$.
In decentralized FL, we hope that the algorithm works with {\em large} $m$ and {\em large} $K$ to support large-scale systems and reduce communication costs, respectively, which in turn leads to a large consensus error.
This large consensus error makes the algorithm harder to converge compared to decentralized learning algorithms.
Therefore, in what follows, we will establish the error bound for the consensus error in Lemma~\ref{Lem: Iterates}. Unlike the conventional gradient-tracking analysis that simply focuses on one iteration (cf., e.g., Lemma 3 in \cite{lu2019gnsd}), our analysis studies the consensus error across {\em multiple inner loop iterations}, which thus is novel and more challenging.
\begin{lem}\label{Lem: Iterates}
Under Assumption~\ref{Assumption: function}, we have the following bounds for the consensus error in Algorithm~\ref{Algorithm: NET-FLEET}:
\begin{align}
 &\!\!\sum_{k=0}^{K-1}\sum_{i=1}^{m}\|\x^{(i)}_{s,k} \!-\! \xb_{s,k}\|^2 
\!\le\!
(1\!+\!\lambda(K\!-\!1))\!\sum_{i=1}^{m}\|\x^{(i)}_{s,0} \!\!-\! \xb_{s,0}\|^2  \!\notag\\
&\quad\quad +\! \frac{\eta^2K^2}{1\!-\!\lambda} \sum_{k=0}^{K-1}\sum_{i=1}^{m}\|\y^{(i)}_{s,k}\! -\! \by_{s,k}\|^2 , \label{Eq: consensus_x}\\
&\!\!\sum_{k=0}^{K-1}\sum_{i=1}^{m}\|\y^{(i)}_{s,k} \!-\! \by_{s,k}\|^2 
\!\le (1\!+\!\lambda(K\!-\!1))\!\sum_{i=1}^{m}\!\|\y^{(i)}_{s,0} 
- \!\by_{s,0}\|^2 \!\!\notag\\
&\quad\quad +\!\frac{24KL^2}{1\!-\!\lambda}\!\sum_{i=1}^{m}\!\|\x^{(i)}_{s,0}\!- \!\xb_{s,0}\|^2 
\!+\!\frac{6mK\sigma^2}{1\!-\!\lambda}
\!+ \!\frac{12\eta^2K^2L^2}{1\!-\!\lambda}\notag\\
&\quad\quad \times \!\sum_{k=0}^{K-1}\sum_{i=1}^{m}\!\|\y^{(i)}_{s,k} \!- \!\by_{s,k}\|^2 \! +\! \frac{12m\eta^2K^2L^2}{1\!-\!\lambda}\sum_{t= 0}^{k-1}\|\by_{s,t}\|^2. \label{Eq: consensus_y} 
\end{align}
\end{lem}

From~\eqref{Eq: consensus_x}-\eqref{Eq: consensus_y}, we can see that the consensus errors on $\x^{(i)}$ and $\y^{(i)}$ are coupled. 
Moreover, the error bounds are accumulated as inner loop rounds $K$ and worker number $m$ increase.
This observation suggests that we need to judiciously design a potential function $\mathfrak{P}_{s,k}$, so that the linear speedup for convergence remains achievable.

By combining Lemmas~\ref{Lem: Descend} and~\ref{Lem: Iterates} and after some algebraic simplifications, we can conclude that:
\begin{align}
&\frac{\eta}{2}\! \sum_{k= 0}^{K\!-\!1}\Eb[\|\nabla f(\xb_{s,k})\|^2] 
\!\le\!
\Eb[\mathfrak{P}_{s,0} \!-\!\mathfrak{P}_{s,K}] 
 \! - \! \frac{\eta C_{\nf}}{2}\sum_{k= 0}^{K\!-\!1}\Eb[\|\bnf_{s,k}\|^2]
 \!\notag\\
 &- \!\frac{C_{\x}}{m^2K}\Eb[\sum_{i=1}^{m}\|\x^{(i)}_{s,0}\!-\!\xb_{s,0}\|^2] 
 \!-\!\frac{C_{\y}C_1\eta^2}{m^2K}\Eb[\sum_{i=1}^{m}\|\y^{(i)}_{s,0}\!-\!\by_{s,0}\|^2] 
 \!\notag\\
 &+\! (\frac{1}{2} \!+
 \!\frac{12C_1LK\eta^2}{(1\!-\!\lambda)m} \!+\! \frac{72LK^2\eta^2}{(1\!-\!\lambda)^2m}) \frac{LK\eta^2\sigma^2}{m} 
\!+\! \frac{36K\eta^2\sigma^2}{(1\!-\!\lambda)^2m}\!+\! \frac{6C_1\eta^2\sigma^2}{(1\!-\!\lambda)mK},\notag
\end{align}
where $C_{\nf}$, $C_{\x}$ and $C_{\y}$ are three constants dependent on the step-size $\eta$ (see detailed definitions in the supplementary material). 
Then, by properly choosing the step-size, we can ensure that $C_{\nf}$, $C_{\x}$ and $C_{\y}$ are positive, and so terms associated with them can be dropped.
Finally, by telescoping the above inequality, we arrive at the desired result as stated in Theorem~\ref{thm:Convergence of NET-FLEET} and the proof is complete.

\section{Experimental Evaluation}\label{sec:numerical}

In this section, we evaluate our NET-FLEET algorithm on MNIST~\cite{lecun2010mnist} and CIFAR-10~\cite{krizhevsky2009learning} datasets.
Our experiments are conducted with four NVIDIA Tesla V100 GPUs.

\smallskip
\textbf{1) Datasets and Learning Models:} 
\textit{1-a) MNIST with Convolutional Neural Networks (CNN):}
We train a CNN classifier on the MNIST~\cite{lecun2010mnist} dataset.
The adopted CNN model has two convolutional layers (size $3 \times 3 \times 16$), each of which is followed by a max-pooling layer with size $2\times 2$ and then a fully connected layer. 
The ReLU activation is used for the two convolutional layers and the ``softmax'' activation is used at the output layer. 
\textit{1-b) CIFAR-10 with Residual Neural Networks (ResNet):} 
We experiment with classification problems over the CIFAR-10~\cite{krizhevsky2009learning} dataset with the ResNet18~\cite{he2016deep} model.
\textit{1-c) Dataset Partition:}
For independent and identically distributed (i.i.d.) data partition, all workers can access the same global dataset; in the case of non-i.i.d. heterogeneous data partition, we use the same data partition strategy as in~\cite{yang2021achieving} that each worker can access data with at most two labels.
Specifally, for the non-i.i.d. setting, we first sort the training data by label, then divide all the training data into 250 shards with 200 data samples, and randomly assign two shards to each client.

\smallskip
\textbf{2) Network System Model:} 
We consider a decentralized network system with $50$ workers.
The network topology $\mathcal{G}$ is generated by the Erd$\ddot{\text{o}}$s-R$\grave{\text{e}}$nyi random graph.
Without specification, we set the edge connectivity probability $p_c=0.5$ for the random graph generation. 
The consensus matrix is chosen as $\W \!= \!\I \!-\! \frac{2\mathbf{L}}{3\lambda_{\text{max}}(\mathbf{L})}$, where $\mathbf{L}$ is the Laplacian matrix of $\mathcal{G}$ and $\lambda_{\text{max}}(\mathbf{L})$ denotes the largest eigenvalue of $\mathbf{L}$.

\smallskip
\textbf{3) Baselines and Parameter Settings:} 
We compare our NET-FLEET algorithm with the state-of-the-art LD-SGD \cite{li2019communication}, GT-SGD \cite{xin2020improved} and DSGD \cite{lian2017can} on decentralized network systems. The number of local update rounds $K$ is set to $10$ for NET-FLEET and LD-SGD.
%
For MNIST on CNN, we choose the initial step-size as 0.01 and reduce the step-size to by half for every 100 iterations. 
The local batch size is fixed at $32$.
For CIFAR-10 on ResNet, we choose the step-size as $0.001$. 
The local batch size is fixed at $128$ for CIFAR-10 training.

\smallskip
\textbf{4) Performance Comparisons:} 

\begin{figure}[b!]
	\centering
	\subfigure[The i.i.d. case.]{
		\includegraphics[width=0.22\textwidth]{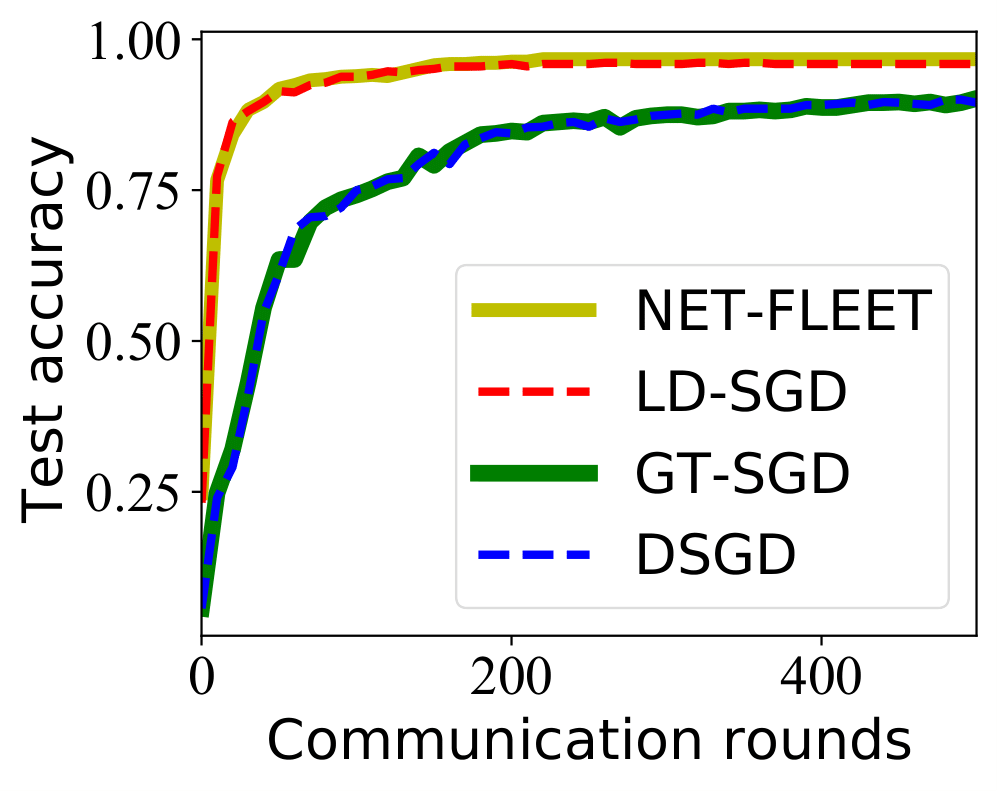}
		\label{fig:CNN_IID_N}
	}
	\subfigure[The non-i.i.d. case.]{
		\includegraphics[width=0.23\textwidth]{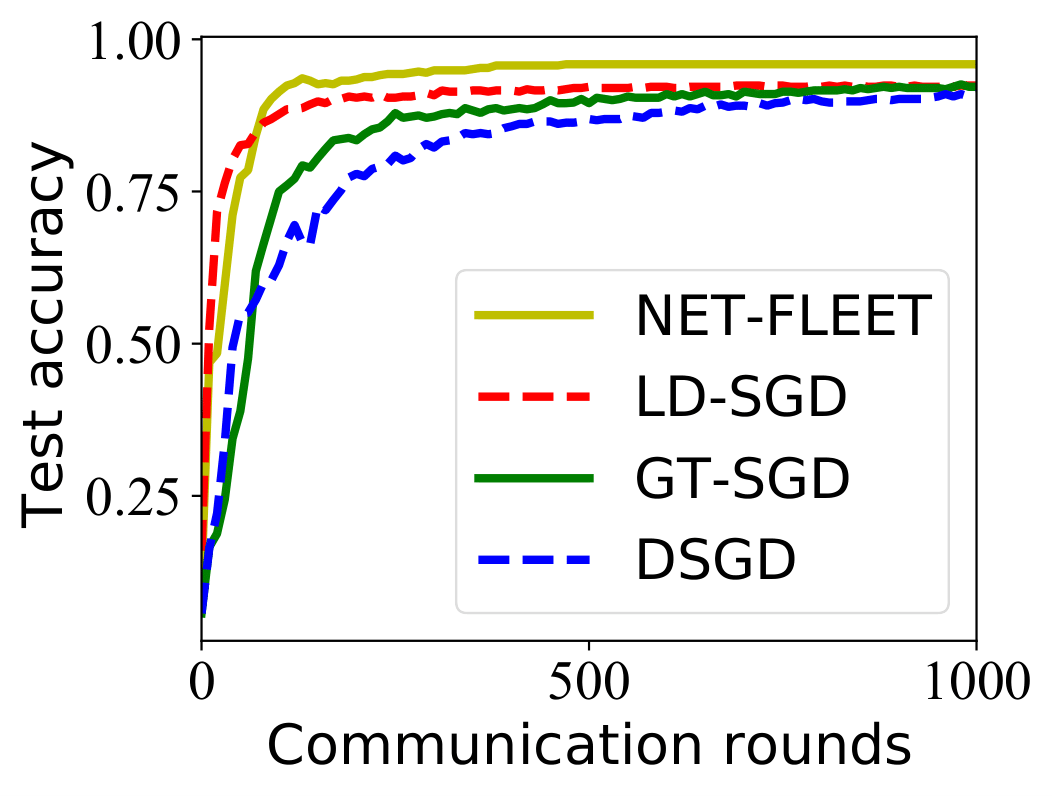}
		\label{fig:CNN_NONIID_N}
	}
	\caption{Test accuracy of CNN on MNIST by different decentralized learning algorithms.}\label{Fig: Decentralized}
\end{figure}

\begin{figure}[b!]
	\centering
	\subfigure[The i.i.d. case.]{
		\includegraphics[width=0.215\textwidth]{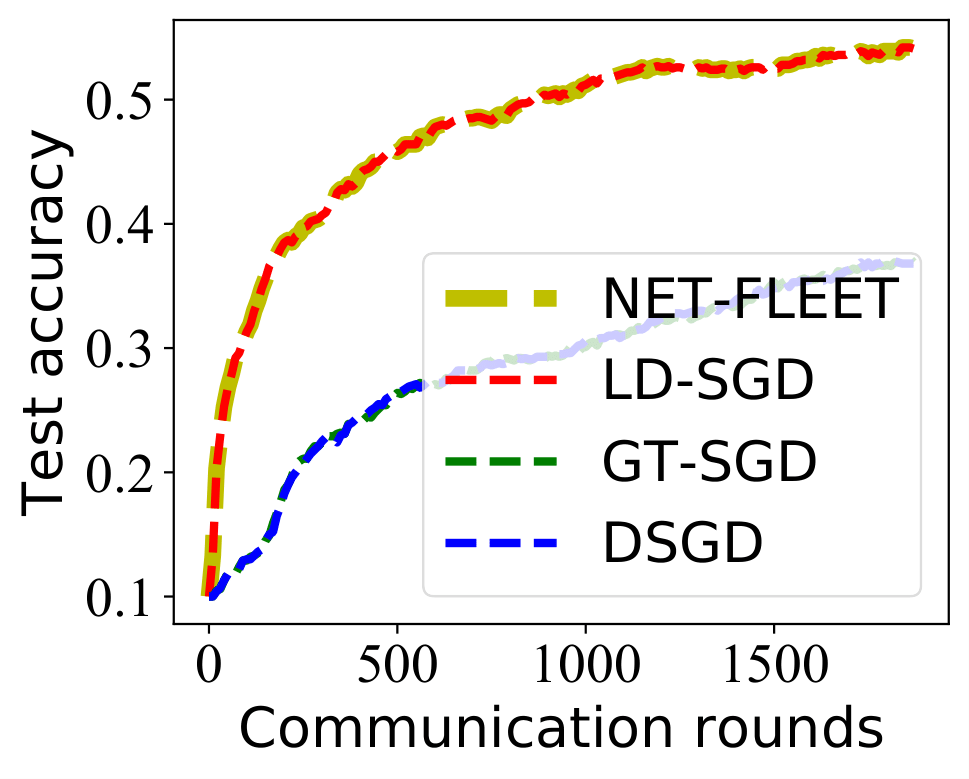}	 
		\label{fig:resnet_IID_N}
	}
	\subfigure[The non-i.i.d. case.]{
		\includegraphics[width=0.215\textwidth]{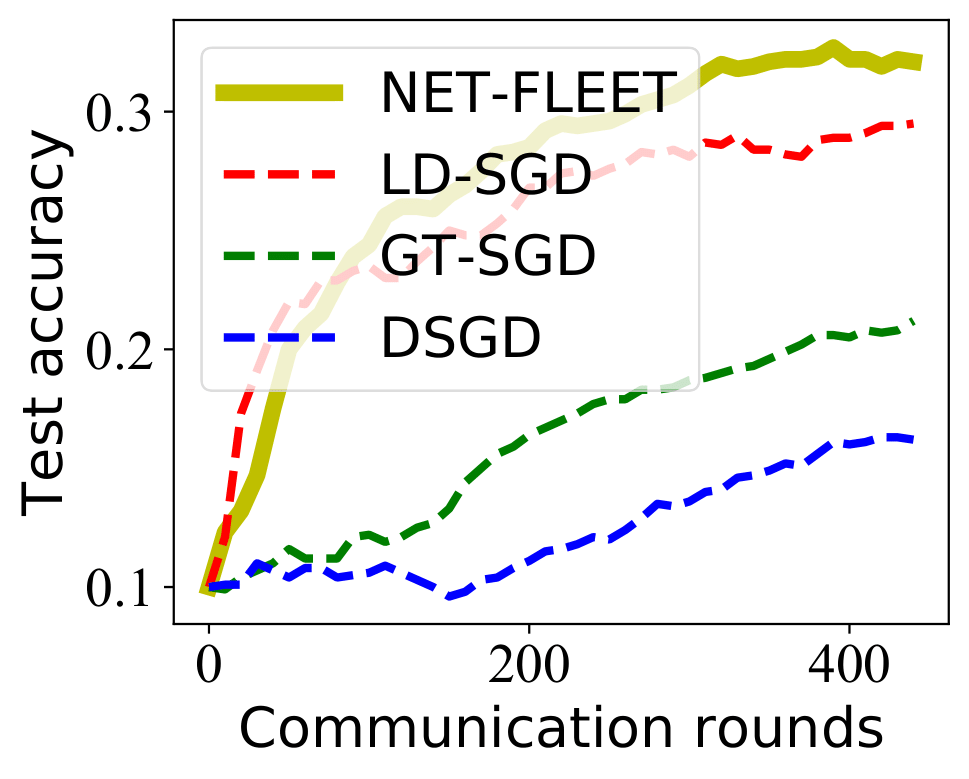}
		\label{fig:resnet_NONIID_N}
	}
	\caption{Test accuracy of ResNet on CIFAR-10 decentralized learning algorithms.}\label{Fig: Decentralized_cifar10_all}
\end{figure}

We compare the test accuracy with respect to the numbers of communication rounds and training samples. 
To better visualize the results, the test accuracies are smoothed by averaging the values in a window of size 10.
Fig.~\ref{Fig: Decentralized} illustrates the results of decentralized algorithms of CNN on MNIST.
In Fig.~\ref{Fig: Decentralized}~(a), we can see that NET-FLEET and LD-SGD have similar performances under i.i.d. data partition and significantly outperform DSGD and GT-SGD with the same communication rounds.
Fig.~\ref{Fig: Decentralized}~(b) shows that under heterogeneous data, NET-FLEET outperforms the other algorithms: with $1000$ communication rounds, the testing accuracy of NET-FLEET is $5\%$ higher than that of LD-SGD and $8\%$ higher than those of DSGD and GT-SGD.

Fig.~\ref{Fig: Decentralized_cifar10_all} illustrates the results of NET-FLEET for ResNet model on CIFAR-10 dataset.
In Fig.~\ref{fig:resnet_IID_N}, we can see that NET-FLEET and LD-SGD have similar performances under i.i.d. data partition and significantly outperform DSGD and GT-SGD with the same number of communication rounds.
Fig.~\ref{fig:resnet_NONIID_N} shows that under heterogeneous data partition, NET-FLEET outperforms the other algorithms: with $500$ communication rounds, the NET-FLEET achieves higher test accuracy than that of LD-SGD, DSGD and GT-SGD.


%

\smallskip
\textbf{5) Impact of the Local Update Rounds:}

\begin{figure}[t!]
	
	\centering
	\subfigure[The i.i.d. case.]{
		\includegraphics[width=0.22\textwidth]{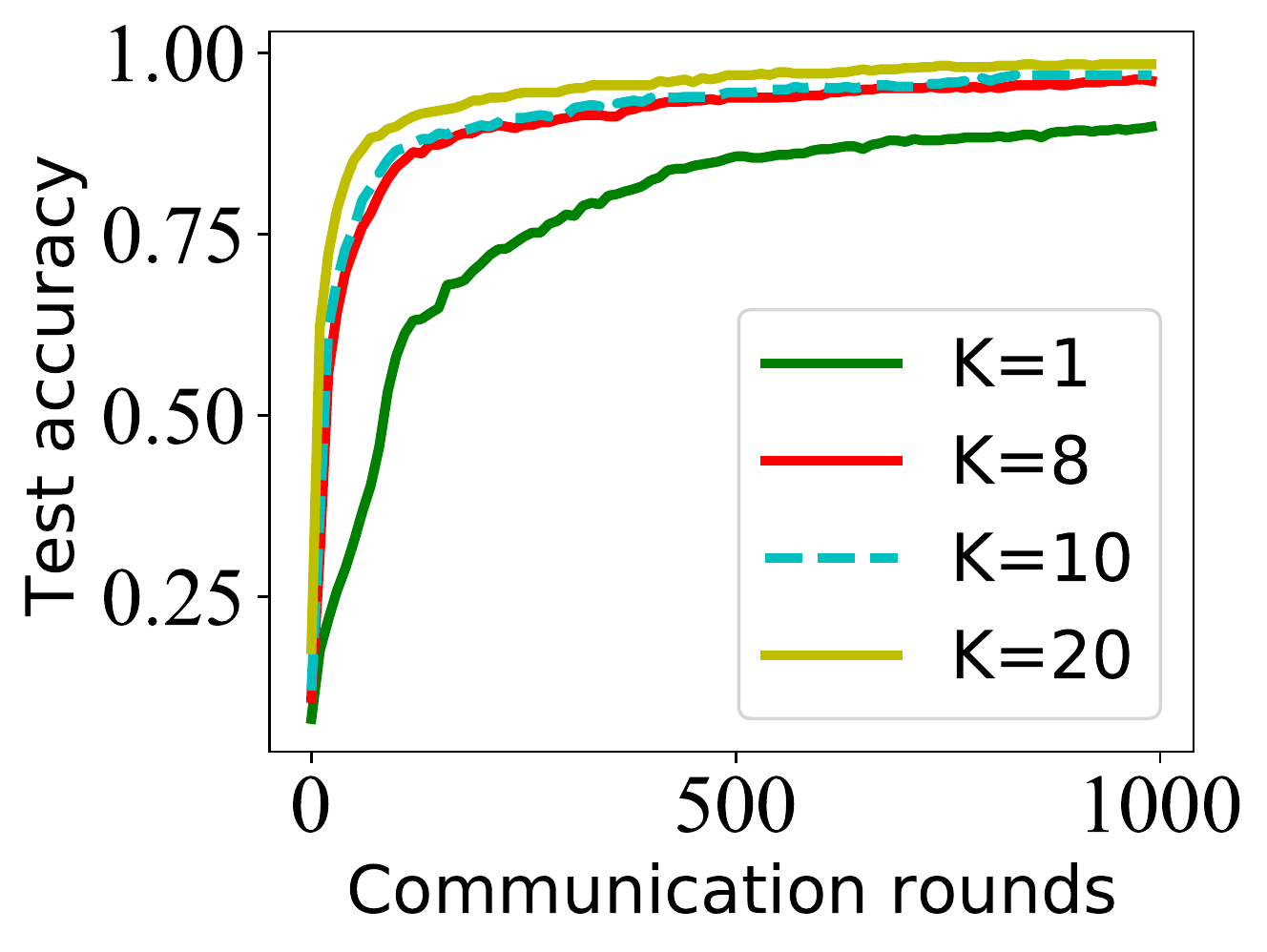}
		\label{fig:small_K_iid}
	}
	\subfigure[The non-i.i.d. case.]{
		\includegraphics[width=0.22\textwidth]{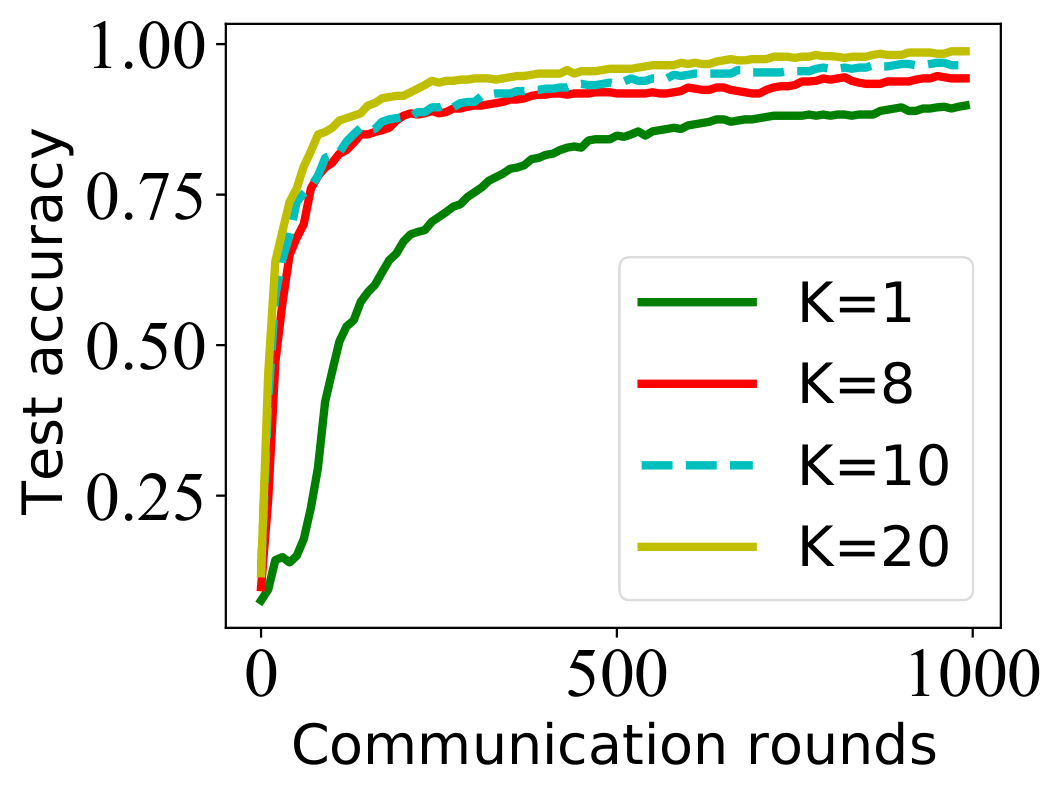}
		\label{fig:small_K_noniid}
	}
	\caption{Test accuracy of CNN on MNIST with different local update rounds.}\label{Fig:K_small}
\end{figure}
%
%

A key feature in FL algorithms is that the workers are allowed to perform multiple local parameter updates.
In this experiment, we examine the impact of different number of local update rounds on the training performance.
We run NET-FLEET to solve classification problems with the CNN model over the MNIST~\cite{lecun2010mnist} dataset. 
We fix the step-size at $0.01$, edge connectivity $p_c$ at $0.5$, local batch size at $32$, and worker number at $50$.
We choose the number of local update rounds $K$ from the discrete set $\{1,8,10,20\}$.
Fig.~\ref{Fig:K_small} shows the performance of NET-FLEET with different number of local update rounds $K$.
As shown in Fig.~\ref{Fig:K_small}, the test accuracy increases as $K$ increases under both the i.i.d. and heterogeneous data settings: with communication rounds being fixed at $1000$, NET-FLEET with $K=1$ has accuracy less than $80\%$.
In contrast, with $K=8$, $K=10$ and $K=20$, NET-FLEET achieves more than $95\%$ testing accuracy.

\begin{figure}[t!]
	
	\centering
	\subfigure[The i.i.d. case.]{
		\includegraphics[width=0.22\textwidth]{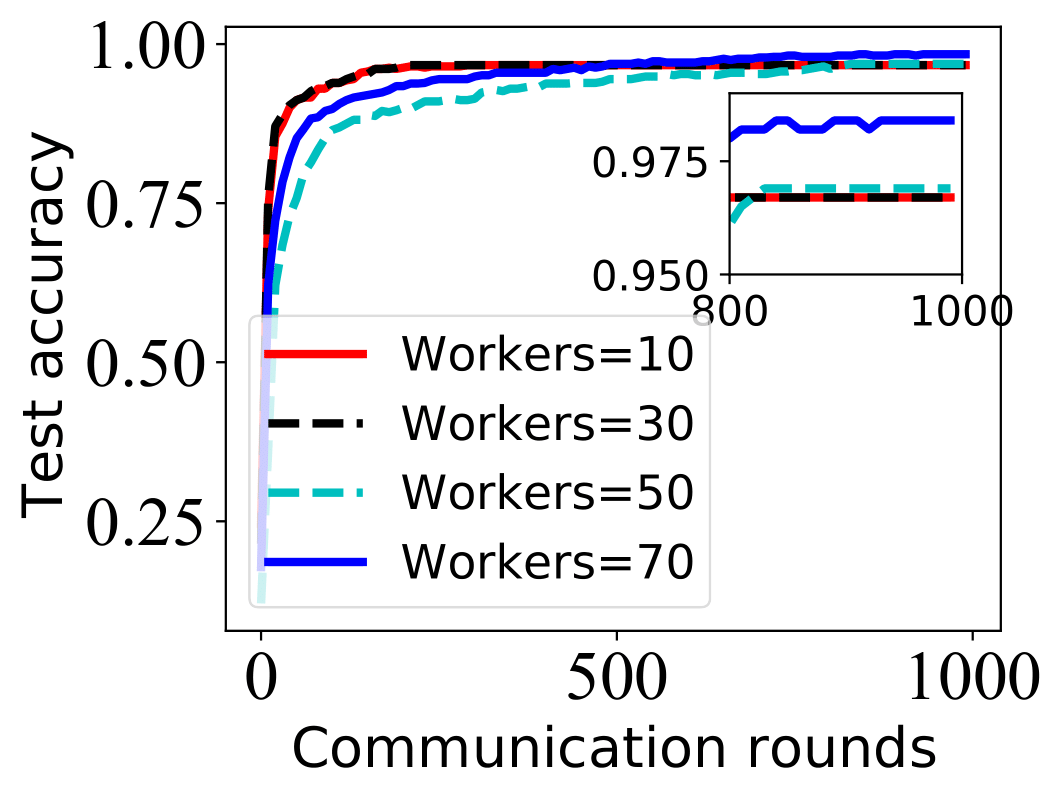}
		\label{fig:worker_iid}
	}
	\subfigure[The non-i.i.d. case.]{
		\includegraphics[width=0.22\textwidth]{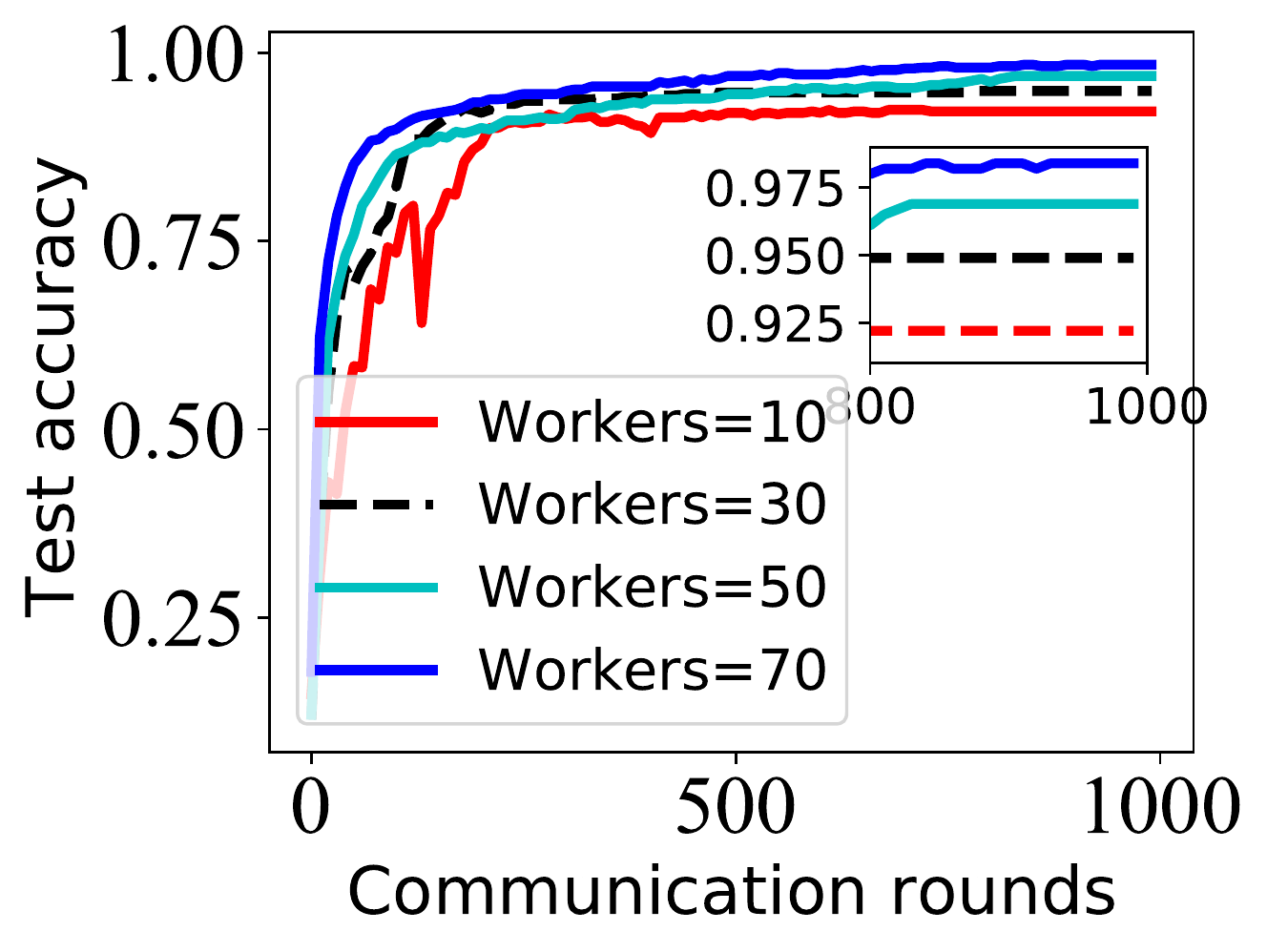}
		\label{fig:worker_noniid}
	}
	\caption{Test accuracy of CNN on MNIST with different number of workers.}\label{Fig:worker}
	\vspace{-0.05in}
\end{figure}

\smallskip
\textbf{6) Impact of the Number of Workers:} We conduct the following experiments with different number of workers.
In this experiment, we choose the number of workers from the discrete set $\{10,30,50,70\}$ and fix the step-size at $0.01$, local update rounds at $10$,  edge connectivity $p_c$ at $0.5$, and local batch size at $32$.
As shown in Fig.~\ref{Fig:worker}, convergence results with different number of workers have similar performances in i.i.d case. 
NET-FLEET achieves $95\%$ accuracy in the i.i.d case.
In the non-i.i.d heterogeneous case, we can see that as the number of workers decreases, the convergence rate decreases.
NET-FLEET obtains an accuracy around $96\%$ with $10$ workers and achieves more than $97.5\%$ test accuracy with $70$ workers in i.i.d case. 
In heterogeneous data case, NET-FLEET's accuracy is approximately $92.5\%$ with $10$ workers and achieves a test accuracy more than $97.5\%$ with $70$ workers.

\smallskip
\textbf{7) Impact of the Edge Connectivity Probability:}

\begin{figure}[t!]
	\centering
	\subfigure[The i.i.d. case.]{
		\includegraphics[width=0.217\textwidth]{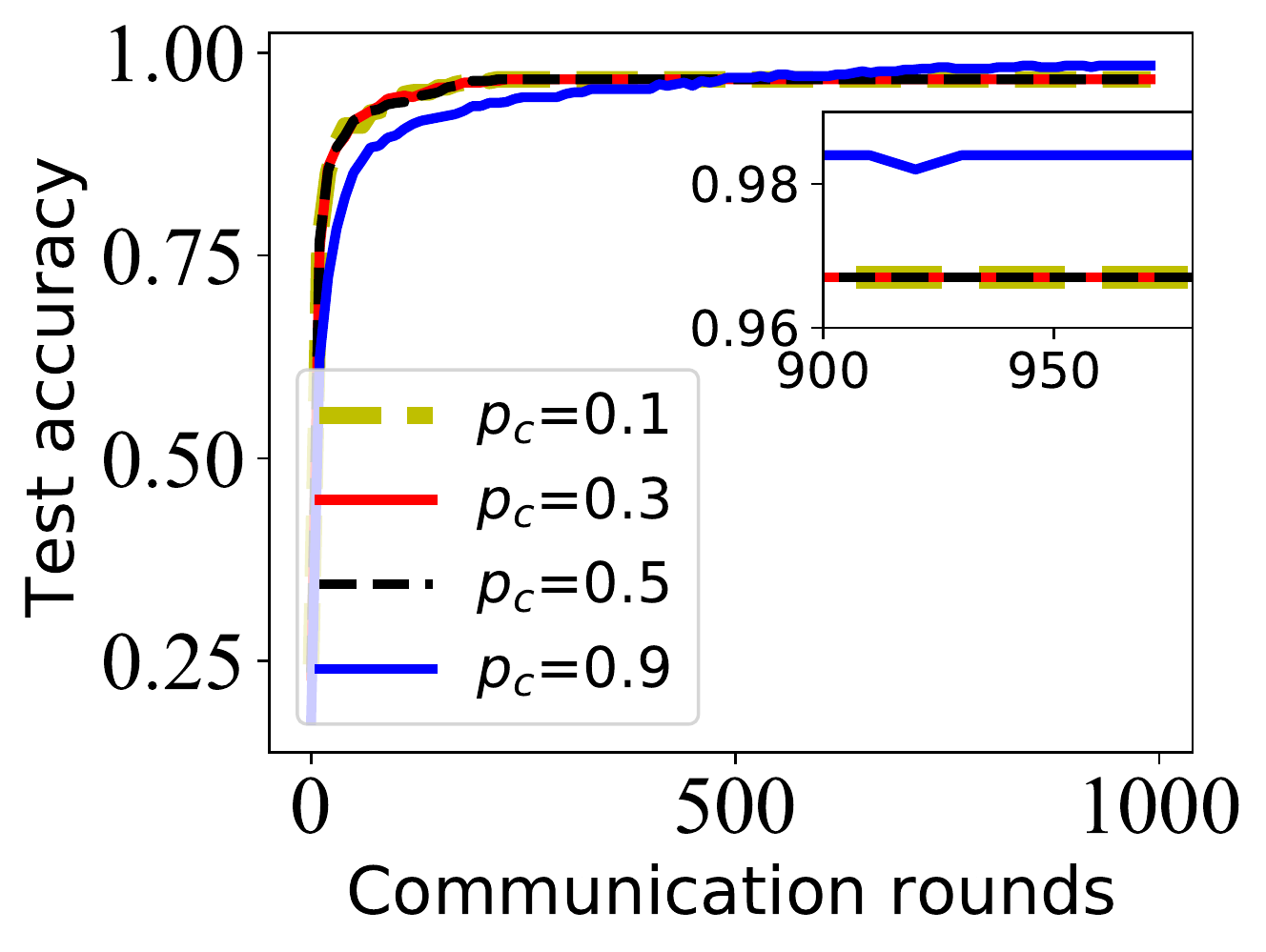}
		\label{fig:probability_iid}
	}
	\subfigure[The non-i.i.d. case.]{
		\includegraphics[width=0.21\textwidth]{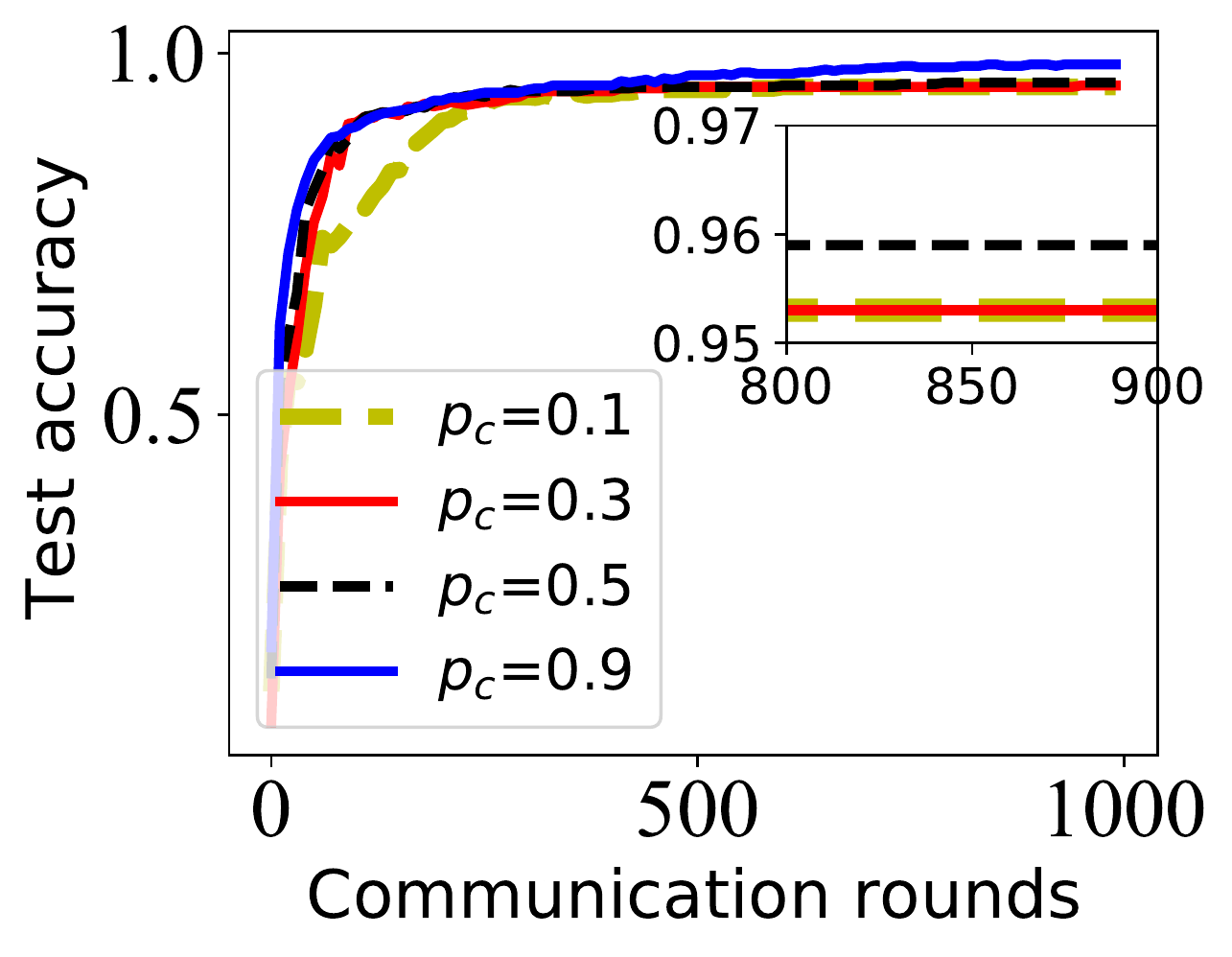}
		\label{fig:probability_noniid}
	}
	\caption{Test accuracy of CNN on MNIST with different edge connection probability $p_c$.}\label{Fig:probability}
\end{figure}

\begin{figure}[t!]
	\centering
	\subfigure[The i.i.d. case.]{
		\includegraphics[width=0.215\textwidth]{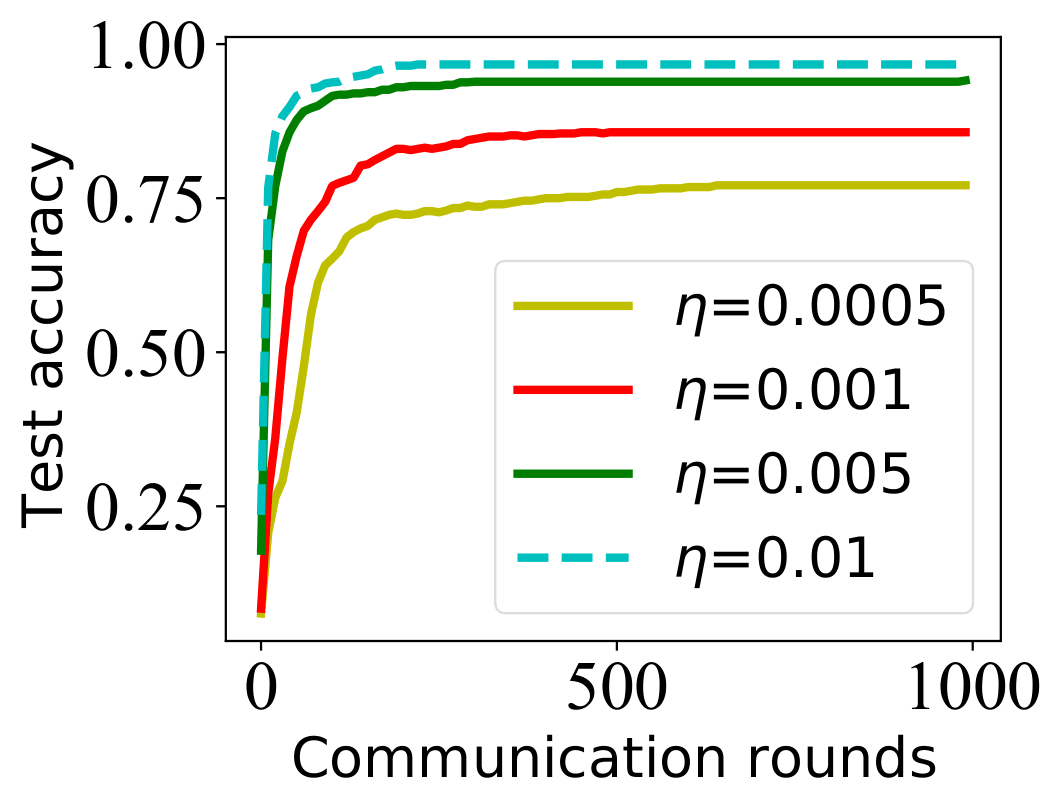}
		\label{fig:step-size_iid}
	}
	\subfigure[The non-i.i.d. case.]{
		\includegraphics[width=0.215\textwidth]{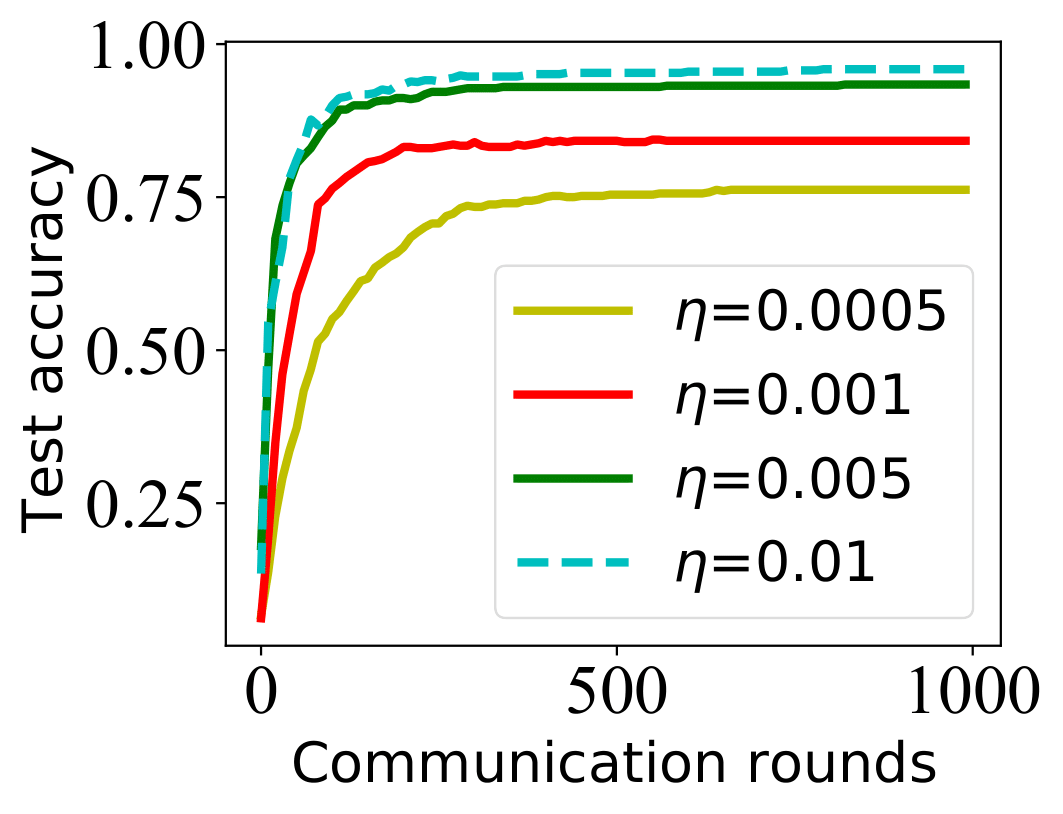}
		\label{fig:step-size_noniid}
	}
	\caption{Test accuracy of CNN on MNIST with different local step-size.}\label{Fig:step-size_appendix}
\end{figure}

For the decentralized network system, the network graph $\mathcal{G}$ is generated by the Erd$\ddot{\text{o}}$s-R$\grave{\text{e}}$nyi random graph with edge connection probability $p_c$.
In the first experiment, we examine the impact of different $p_c$-values on the training performance with the CNN model over the MNIST dataset. We choose the $p_c$-value from the discrete set $\{0.1,0.3,0.5,0.9\}$ and fix the number of workers at $50$, local update rounds at $10$, step-size at $0.01$, and local batch size at $32$.
Fig.~\ref{Fig:probability} shows that the convergence result with different edge connectivity $p_c$-values have similar performances in the i.i.d case. 
The experiments achieve a $96\%$ accuracy in the i.i.d case.
In the heterogeneous data case, we can see that as $p_c$ increases, the test accuracy increases slightly, which shows that the learning performance of NET-FLEET is insensitive to the $p_c$-value.

\textbf{8) Impact of the Step-size:} 
In this experiment, we choose the step-size from the discrete set $\{0.0005, 0.001, 0.005, 0.01\}$ and fix worker number at $50$, local update rounds at $10$,  edge connectivity $p_c$  at $0.5$, and local batch size at $32$, global batch size at $512$.
As shown in Fig.~\ref{Fig:step-size_appendix}, larger local step-sizes lead to faster convergence rates in both i.i.d and non-i.i.d cases.
NET-FLEET achieves accuracy less than $75\%$ with a step-size $0.0005$, and obtains more than $95\%$ test accuracy with a step-size $0.01$.

\section{Conclusion}\label{sec:conclusion}

In this paper, we studied fully decentralized federated learning with data heterogeneity.
A novel federated learning algorithm named NET-FLEET was proposed for fully decentralized network systems.
Our NET-FLEET algorithm allows the workers to keep the local data and run multiple local update steps during the training, thus maintaining local data privacy and reducing the communication costs. 
We showed that with properly selected parameters, our algorithm achieves the state-of-the-art linear speedup for convergence, i.e., an $O(1/\sqrt{mKS})$ convergence rate, where $m$ is the number of workers, and $S$ and $K$ are the numbers of communication and local update rounds, respectively.
Extensive numerical studies verified the theoretical performance results of our proposed algorithm.

\section*{Acknowledgments}

This work has been supported in part by NSF grants CAREER CNS-2110259, CNS-2112471, CNS-2102233, CCF-2110252, CCF 1934884, and SES 1952007.

\bibliographystyle{acm}
\bibliography{reference.bib}

  \appendix

\section{Proof of Main Results}

For notation convenience, we define the following variables: $\widetilde{\W} = \W \otimes \I_m$, $\g_{s,k}^{(i)} = \nabla f_i(\x_{s,k}^{(i)};\vzeta_{s,k}^{(i)})$, $\nf_{s,k}^{(i)} = \nabla f_i(\x_{s,k}^{(i)})$,  and $\a_{s,k} = [\a_{s,k}^{(i)\top},\cdots,\a_{s,k}^{(i)\top}]^\top$ and $\bar{\a}_{s,k}=\frac{1}{m} \sum_{i=1}^{m}\a_{s,k}^{(i)},$ for $\a \in \{\x,\y, \g,\nf\}$.
Here $\by_{s,k} = \bg_{s,k}$ because of $\by_{s,0} =  \bg_{s,0}.$ 
Also, we define matrix $\Q \triangleq \I - (\frac{1}{m}\1\1^\top )\otimes \I$, so it holds that $\Q\a_{s,k} = \a_{s,k} - \1\otimes \bar{\a}_{s,k}$.


\subsection{Proof of Lemma~\ref{Lem: Descend}}

\begin{proof}
From the $L$-smoothness of $f$ and $\xb_{s,k+1}= \xb_{s,k} - \eta \by_{s,k} = \xb_{k} - \eta \bg_{s,k},$ we have
\begin{align}
&f(\xb_{s,k+1})
\!\le\!
f(\xb_{s,k}) -  \langle \nabla f(\xb_{s,k}) , \xb_{s,k+1}- \xb_{s,k} \rangle + \frac{L}{2}\|\xb_{s,k+1}\notag\\
&- \xb_{s,k}\|^2 
=
f(\xb_{s,k}) - \eta \langle \nabla f(\xb_{s,k}) , \bg_{s,k} \rangle + \frac{L\eta^2}{2}\|\bg_{s,k}\|^2. 
\end{align}
Since $\Eb[\g_{s,k}^{(i)}| \Fc_{s,k}] = \nabla f_{s,k}^{(i)}$, we have 
\begin{align}
&
\Eb[f(\xb_{s,k\!+\!1})| \Fc_{s,k}] 
\!\le\!
f(\xb_{s,k}) \!-\! \eta \langle \nabla f(\xb_{s,k}) , \!\bnf_{s,k} \rangle \!+\! \frac{L\eta^2}{2}\Eb[\|\bg_{s,k}\|^2  \notag\\
&
| \Fc_{s,k}]
=
f(\xb_{s,k}) - \frac{\eta}{2}\|\nabla f(\xb_{s,k})\|^2 - \frac{\eta}{2} \|\bnf_{s,k}\|^2 + \frac{\eta}{2}\|\nabla f(\xb_{s,k})\notag\\
&
-\bnf_{s,k}\|^2 + \frac{L\eta^2}{2}\Eb[\|\bg_{s,k}\|^2| \Fc_{s,k}] 
\le
f(\xb_{s,k}) - \frac{\eta}{2}\|\nabla f(\xb_{s,k})\|^2 \notag\\
&- \frac{\eta}{2} \|\bnf_{s,k}\|^2 + \frac{L^2\eta}{2m}\|\Q\x_{s,k}\|^2 + \frac{L\eta^2}{2}\Eb[\|\bg_{s,k}\|^2| \Fc_{s,k}]. 
\end{align}

Taking full expectation on the above inequality and telescoping from $k = 0$ to $K-1$ yields:
\begin{align}
&\Eb[f(\xb_{s,K}) \!-\!f(\xb_{s,0})]
\!\le \!
\!-\! \frac{\eta}{2} \!\sum_{k= 0}^{K-1}\Eb[\|\nabla f(\xb_{s,k})\|^2] 
 \!- \!\frac{\eta}{2} \!\sum_{k= 0}^{K-1}\Eb[\|\bnf_{s,k}\|^2] \notag\\
 &
 \!+\! \frac{L\eta^2}{2}\! \sum_{k= 0}^{K-1}\Eb[\|\bg_{s,k}\|^2] 
 \!+\! \frac{L^2\eta}{2m} \!\sum_{k= 0}^{K-1}\Eb[\|\Q\x_{s,k}\|^2 ].
\end{align}
\end{proof}

\subsection{Proof of Lemma~\ref{Lem: Iterates}}

\begin{proof}
First, for any $\x_t$ and $\lambda = \max\{|\lambda_2|,|\lambda_m|\}$, we have:
\begin{align}\label{Eq: contra}
\|\widetilde{\W}\x_{t} \!-\!\1\otimes\xb_{t} \|^2 \!=\! \|\widetilde{\W}(\x_{t} \!-\!\1\otimes\xb_{t}) \|^2 \!\le\! \lambda^2\|\x_{t} \!-\!\1\otimes\xb_{t}\|^2.
\end{align}

Note that $\x_{s,k} = \widetilde{\W} \x_{s,0} - \eta \sum_{t=0}^{k-1} \y_{s,t}$ and $\xb_{s,k} = \xb_{s,0} - \eta \sum_{t=0}^{k-1} \by_{s,t}$.
Thus, we have 
\begin{align}\label{Eq: Qx}
&
\|\Q\x_{s,k}\|^2 
=
\|\widetilde{\W} \x_{s,0} - \eta \sum_{t=0}^{k-1} \y_{s,t}-\1\otimes(\xb_{s,0} - \eta \sum_{t=0}^{k-1} \by_{s,t})\|^2 \notag\\
&
\stackrel{(a)}{\le}
(1\!+\!c_1)\|\widetilde{\W}\x_{s,0} \!-\!\1\!\otimes\!\xb_{s,0} \|^2 
\!+\! 
(1\!+\!\frac{1}{c_1}) \eta^2\|\sum_{t=0}^{k-1}\y_{s,t}\!-\!\1\!\otimes\! \by_{s,t}\|^2 \notag\\
&
\stackrel{(b)}{\le}
\lambda\|\x_{s,0} -\1\otimes\xb_{s,0} \|^2 
+ 
\frac{\eta^2}{1-\lambda}\|\sum_{t=0}^{k-1}\y_{s,t}-\1\otimes \by_{s,t}\|^2 \notag\\
&
\stackrel{(c)}{\le}
\lambda\|\Q\x_{s,0}\|^2 + \frac{\eta^2k}{1-\lambda} \sum_{t=0}^{k-1}\|\Q\y_{s,t}\|^2,  
\end{align}
where (a) follows from $\|\x+\y\|^2 \le (1+c) \|\x\|^2 + (1+1/c)\|\y\|^2$ for any $c > 0$, (b) follows from~\eqref{Eq: contra} with $c_1 = 1/\lambda-1$, and (c) follows from the Jensen's inequality.

Since $\y_{s,k}\! =\!   \y_{s,k-1}  \!+ \g_{s,k}\!- \g_{s,k-1} \!=\! \widetilde{\W} \y_{s,0} \!+\! \g_{s,k}\!-\! \g_{s,0}$ and $\by_{s,k}\! =  \!\by_{s,0} \!+ \bg_{s,k}\!- \bg_{s,0}$, it follows that 
\begin{align}\label{Eq: Qy}
&
\|\Q\y_{s,k}\|^2
=
\|\widetilde{\W} \y_{s,0} + \g_{s,k}- \g_{s,0} -\1\otimes(\by_{s,0} + \bg_{s,k}- \bg_{s,0} )\|^2 \notag\\
&
{\le}
\lambda\|\Q\y_{s,0}  \|^2 
+ 
\frac{1}{1-\lambda} \|\g_{s,k}- \g_{s,0}\|^2 \notag\\
&{\le}
\lambda\|\Q\y_{s,0}  \|^2 
+ 
\frac{3}{1-\lambda}(2m\sigma^2+L^2\|\x_{s,k}- \x_{s,0}\|^2),
\end{align}

Note that the term $\|\x_{s,k}- \x_{s,0}\|^2 $ can be bounded as:
\begin{align}\label{Eq: x-x}
&
\|\x_{s,k}\!- \!\x_{s,0}\|^2 
\!= \!
\|\widetilde{\W} \x_{s,0}\!- \! \eta \!\sum_{t=0}^{k-1} \y_{s,t} \!-\! \x_{s,0}\|^2 
\!=\!
8\| \x_{s,0} \!-\! \1\!\otimes\!\xb_{s,0} \|^2 \notag\\
&
\!+ \!2\eta^2k\sum_{t=0}^{k-1} \| \y_{s,t}\|^2 
\stackrel{(a)}{\le}
8\| \x_{s,0} \!- \!\1\!\otimes\!\xb_{s,0} \|^2 \!+\! 2\eta^2k\!\sum_{t=0}^{k-1} (2\|\1\!\otimes\! \by_{s,t}\|^2\notag\\
& \!+\!2\| \y_{s,t} \!- \!\1\!\otimes\! \by_{s,t}\|^2)
\!\le\!
8\|\Q\x_{s,0} \|^2 \!+\! 4\eta^2k\!\sum_{t=0}^{k\!-\!1} \| \Q\y_{s,t}\|^2 \notag\\
&\!+\! 4\eta^2mk\!\sum_{t=0}^{k\!-\!1} \| \by_{s,t}\|^2,
\end{align}
where $(a)$ is due to the fact that $\|\widetilde{\W} -\I\| \le 2.$

Thus, by plugging~\eqref{Eq: x-x} into~\eqref{Eq: Qy}, we have  
\begin{align}
&\|\Q\y_{s,k}\|^2 
\!\le\!
\lambda\|\Q\y_{s,0} \|^2 
+ \frac{6m\sigma^2}{1-\lambda}
+
\frac{24L^2}{1-\lambda}\|\Q\x_{s,0}\|^2 \notag\\
&+ \frac{12k\eta^2L^2}{1-\lambda} \sum_{t=0}^{k-1}\|\Q\y_{s,t}\|^2 + \frac{12mk\eta^2L^2}{1-\lambda}\sum_{t= 0}^{k-1}\|\by_{s,t}\|^2.
\end{align}

\end{proof}

\subsection{Proof of Theorem~\ref{thm:Convergence of NET-FLEET}}
\begin{proof}
By combining the results from Lemma~\ref{Lem: Descend} and Lemma~\ref{Lem: Iterates}, we have 
\begin{align}\label{Eq: diff_f}
&\Eb[f(\xb_{s,K}) \!-\!f(\xb_{s,0})]
\!\le\! 
-\! \frac{\eta}{2} \sum_{k= 0}^{K-1}\Eb[\|\nabla f(\xb_{s,k})\|^2] 
 \!-\! \frac{\eta}{2} \sum_{k= 0}^{K-1}\Eb[\|\bnf_{s,k}\|^2]\notag\\
 & \!- \!\frac{L^2\eta}{2m} \sum_{k= 0}^{K-1}\Eb[\|\Q\x_{s,k}\|^2 ]
 \!+\! \frac{L\eta^2}{2} \sum_{k= 0}^{K-1}\Eb[\|\bg_{s,k}\|^2]  
 \!+\! \frac{L^2\eta}{m}(1+\lambda(K-1))\notag\\
 &\times \Eb\|\Q\x_{s,0}\|^2 \!+\! \frac{\eta^3L^2K^2}{m(1-\lambda)} \sum_{k=0}^{K-1}\Eb\|\Q\y_{s,k}\|^2.
\end{align}

Also, from Lemma~\ref{Lem: Iterates}, for some constant $C_1$ (to be determined later), it follows that 
\begin{align}\label{Eq: diff_p}
&(\|\Q\x_{s,K}\|^2 + C_1\eta^2\|\Q\y_{s,K}\|^2) - 
(\|\Q\x_{s,0}\|^2 + C_1\eta^2\|\Q\y_{s,0}\|^2) \notag\\
\le
&
-(1-\lambda - \frac{24C_1L^2\eta^2}{1-\lambda})\|\Q\x_{s,0}\|^2
-(1-\lambda)C_1\eta^2\|\Q\y_{s,0}\|^2 \notag\\
&
+\! \frac{\eta^2K \!+ \!12 C_1KL^2\eta^4}{1\!-\!\lambda} \sum_{k=0}^{K-1}\|\Q\y_{s,k}\|^2 
\!+\! \frac{12mC_1L^2K^2\eta^4}{1\!-\!\lambda}\notag\\
&
\times \sum_{k=0}^{K-1}\|\by_{s,k}\|^2 
\!+\! \frac{6mC_1\eta^2\sigma^2}{1\!-\!\lambda}.
\end{align}

Thus, combining~\eqref{Eq: diff_f} and~\eqref{Eq: diff_p}, we have
\begin{align}\label{Eq: Desend_fp}
&\Eb[f(\xb_{s,K}) \!-\!f(\xb_{s,0}) \!+\! \frac{1}{m^2K}(\|\Q\x_{s,K}\|^2 \!+\! C_1\eta^2\|\Q\x_{s,K}\|^2)\notag\\
& \!-\! 
\frac{1}{m^2K}(\|\Q\x_{s,0}\|^2 \!+\! C_1\eta^2\|\Q\x_{s,0}\|^2)]\notag\\
\!\le\! 
&
-\! \frac{\eta}{2} \!\sum_{k= 0}^{K-1}\!\!\Eb[\|\nabla f(\xb_{s,k})\|^2] 
 \!-\! \frac{\eta}{2} \!\!\sum_{k= 0}^{K-1}\!\!\Eb[\|\bnf_{s,k}\|^2]\!- \!\frac{L^2\eta}{2m}\!\! \sum_{k= 0}^{K-1}\!\!\Eb[\|\Q\x_{s,k}\|^2 ]\notag\\
 &
\!+\! \frac{6C_1\eta^2\sigma^2}{(1\!-\!\lambda)mK}
 \!+\! (\frac{L\eta^2}{2}\!+\!\frac{12C_1L^2K\eta^4}{(1\!-\!\lambda)m}) \sum_{k= 0}^{K-1}\!\!\Eb[\|\bg_{s,k}\|^2]  
 \!+ \!\frac{3\eta^2}{(1\!-\!\lambda)m^2} \notag\\
 &
 \times\!\sum_{k=0}^{K-1}\!\!\Eb\|\Q\y_{s,k}\|^2
 \!-\!(1\!-\!\lambda\! -\! \frac{24C_1L^2\eta^2}{1\!-\!\lambda}\!- \!(1\!+\!\lambda(K-1))mKL^2\eta)\notag\\
 &
 \times \frac{1}{m^2K}\Eb[\|\Q\x_{s,0}\|^2]-(1-\lambda)\frac{C_1\eta^2}{m^2K}\Eb[\|\Q\y_{s,0}\|^2] ,
\end{align}
by setting $\eta \le \min\{1/mL^2K^2, 1/\sqrt{12C_1L^2}\}$.

From Lemma~\ref{Lem: Iterates}, with $\eta \le \sqrt{\frac{1-\lambda}{24L^2K^2}}$, it holds that
\begin{align}\label{Eq: sum_Qy2}
&\sum_{k=0}^{K-1}\|\Q\y_{s,k}\|^2 
\le
2(1+\lambda(K-1))\|\Q\y_{s,0} \|^2 
+ \frac{12mK\sigma^2}{1-\lambda}\notag\\
&
+
\frac{48KL^2}{1-\lambda}\|\Q\x_{s,0}\|^2 
+ \frac{24m\eta^2K^2L^2}{1-\lambda}\sum_{t= 0}^{k-1}\|\by_{s,t}\|^2.
\end{align}

By plugging~\eqref{Eq: sum_Qy2} into~\eqref{Eq: Desend_fp}, we have 
\begin{align}
&\Eb[f(\xb_{s,K}) \!-\!\!f(\xb_{s,0}) \!+\!\! \frac{1}{mK}(\|\Q\x_{s,K}\|^2 \!\!\!+\!\! C_1\eta^2\|\Q\x_{s,K}\|^2) \!-\! \!
\frac{1}{mK}(\|\Q\x_{s,0}\|^2\notag\\&\! +\! C_1\eta^2\|\Q\x_{s,0}\|^2)]
\!\le\!
\!-\! \frac{\eta}{2} \sum_{k= 0}^{K-1}\!\!\Eb[\|\nabla f(\xb_{s,k})\|^2] 
\!-\! \frac{L^2\eta}{2m} \sum_{k= 0}^{K-1}\!\!\Eb[\|\Q\x_{s,k}\|^2 ]\notag\\
 &
  \!-\! \frac{\eta C_{\nf} }{2}\sum_{k= 0}^{K-1}\!\!\Eb[\|\bnf_{s,k}\|^2]
 \!-\!\frac{C_{\x}}{m^2K}\!\!\Eb[\|\Q\x_{s,0}\|^2] 
 \!-\!\frac{C_{\y}C_1\eta^2}{m^2K}\!\!\Eb[\|\Q\y_{s,0}\|^2]  \notag\\
 &
 \!+\! (\frac{L\eta^2}{2}\!+\!\frac{12C_1L^2K\eta^4}{(1\!-\!\lambda)m} \!+\! \frac{72L^2K^2\eta^4}{(1-\lambda)^2m}) \frac{K\sigma^2}{m}  
 \!+\! \frac{36K\eta^2\sigma^2}{(1\!-\!\lambda)^2m}\!+\! \frac{6C_1\eta^2\sigma^2}{(1\!-\!\lambda)mK}.
\end{align}
where
$C_{\nf} \triangleq 1- L\eta-\frac{24C_1L^2K\eta^3}{(1\!-\!\lambda)m} - \frac{144L^2K^2\eta^3}{(1-\lambda)^2m}$, $C_{\x} \triangleq 1-\lambda - \frac{24C_1L^2\eta^2}{1-\lambda} - (1+\lambda(K-1))mKL^2\eta - \frac{144L^2\eta^2K^2}{(1-\lambda)^2}$, 
$C_{\y} \triangleq 1-\lambda - \frac{6(1+\lambda K -\lambda)K}{C_1(1-\lambda)}$.
By setting $C_1 = \frac{6(1+\lambda K-\lambda)K}{(1-\lambda)^2}$, we have $C_{\y} = 0$. 
By letting $\eta \le \min\{\sqrt{\frac{m(1-\lambda)^3}{144(1+\lambda K -\lambda)LK^2}},\sqrt{\frac{m(1-\lambda)^2}{144LK^2}}, 1/3L\}$, we have $C_{\nf} \ge 0$.
Also, letting $\eta \le \min\{\frac{(1-\lambda)}{3(1+\lambda K -\lambda)mKL^2}, \frac{(1-\lambda)^3mK}{144}, \frac{(1-\lambda)^2(1+\lambda K -\lambda)m}{144K}\}$, we have $C_{\x} \ge 0$.

With the above parameter setting and the proposed potential function, we have 
\begin{align}\label{Eq: error_eq}
&\frac{\eta}{2} \sum_{k= 0}^{K-1}\Eb[\|\nabla f(\xb_{s,k})\|^2 +\frac{L^2}{m}\|\Q\x_{s,k}\|^2]  
 \stackrel{}{\le} 
 \Eb[\mathfrak{P}_{s,0}\!-\!\mathfrak{P}_{s,K}]\notag\\&   \!+\! \frac{3LK\sigma^2\eta^2}{2m}  
 \!+\! \frac{36K\eta^2\sigma^2}{(1\!-\!\lambda)^2m}\!+\! \frac{36(1+\lambda K \!-\!\lambda )\eta^2\sigma^2}{(1\!-\!\lambda)^3m} ,
\end{align}
by further setting $\eta \le \min\{\sqrt{\frac{m(1-\lambda)^3}{144(1+\lambda K -\lambda)LK^2}}, \sqrt{\frac{m(1-\lambda)^2}{144LK^2}}\}$.

Telescoping~\eqref{Eq: error_eq} for $s$ from $0$ to $S-1$ and multiplying the factor $2SK/\eta$ on both sides, we have 
\begin{align}
&\frac{1}{SK} \sum_{s\!=\! 0}^{S\!-\!1}\sum_{k\!=\! 0}^{K\!-\!1}\Eb[\|\nabla f(\xb_{s,k})\|^2 \!+\!\frac{L^2}{m}\|\Q\x_{s,k}\|^2]  
\!\le\!   \frac{2\Eb[\mathfrak{P}_{0,0}\!-\!\mathfrak{P}_{S,0}]}{SK\eta}\notag\\
&\! + \!\frac{3L\sigma^2\eta}{m}  
 \!+ \!\frac{72\eta\sigma^2}{(1\!-\!\lambda)^2m}\!+\! \frac{72(1+\lambda K -\lambda )\eta\sigma^2}{(1\!-\!\lambda)^3Km} 
\end{align}
This completes the proof of Theorem~\ref{thm:Convergence of NET-FLEET}.
\end{proof}

\subsection{Proof of Corollary~\ref{cor: linear speedup}}
\begin{proof}
Recall from Theorem~\ref{thm:Convergence of NET-FLEET} that the condition on the step-size is
\begin{align}
\eta \!\le 
\!\min\{
&
\frac{1}{3L},
\underbrace{\frac{1}{mL^2K^2}}_{\triangleq r_1}, 
\underbrace{ \frac{(1-\lambda)}{\sqrt{12(1+\lambda K -\lambda)KL^2}}}_{\triangleq r_2},
\underbrace{\sqrt{\frac{1-\lambda}{24L^2K^2}}}_{\triangleq r_3} ,
\notag\\
& 
\underbrace{\sqrt{\frac{m(1-\lambda)^2}{144LK^2}}}_{\triangleq r_5}, 
\underbrace{\frac{(1-\lambda)}{3(1+\lambda K -\lambda)mKL^2}}_{\triangleq r_6}, \notag\\
&
\underbrace{\frac{(1-\lambda)^3mK}{144}}_{\triangleq r_7}, 
\underbrace{\frac{(1-\lambda)^2(1+\lambda K -\lambda)m}{144K}}_{\triangleq r_8}
\}.
\end{align}
where (a) follows from plugging $C_1= 6(1+\lambda K - \lambda)K/(1-\lambda)^2$, and (b) is due to $r_4\le r_5$.

Setting $K = \sqrt[4]{SK/m^3}$ (i.e. $K = S^{1/3}/m$), we have 
\begin{align}
r_1 & = \frac{1}{mL^2K^2} = \frac{\sqrt{m}}{L^2\sqrt{SK}} = O(\frac{\sqrt{m}}{\sqrt{SK}}),\notag\\
r_2 & = \frac{(1\!-\!\lambda)}{\sqrt{12(1\!+\!\lambda K \!-\!\lambda)KL^2}} \!\stackrel{(a)}{\ge}\! \frac{(1\!-\!\lambda)}{\sqrt{12 }K L} \!=\! \frac{(1\!-\!\lambda)m^{3/4}}{\sqrt{12}L(SK)^{1/4}} \!>\! O(\frac{\sqrt{m}}{\sqrt{SK}}),\notag\\
r_3 & = \sqrt{\frac{1-\lambda}{24L^2K^2}} = \sqrt{\frac{1-\lambda}{24L^2}}\frac{m^{3/4}}{(SK)^{1/4}}> O(\frac{\sqrt{m}}{\sqrt{SK}}) \notag\\
r_5 & = \sqrt{\frac{m(1-\lambda)^2}{144LK^2}} = \sqrt{\frac{(1-\lambda)^2}{144L}}\frac{m^{7/4}}{(SK)^{1/4}}> O(\frac{\sqrt{m}}{\sqrt{SK}}) \notag\\
r_6 & = \frac{(1-\lambda)}{3(1+\lambda K -\lambda)mKL^2} \ge \frac{(1-\lambda)}{3mK^2L^2} = \frac{(1-\lambda)\sqrt{m}}{3L^2\sqrt{SK}} = O(\frac{\sqrt{m}}{\sqrt{SK}}) \notag\\
r_7 & = O((SKm)^{1/4})\stackrel{(b)}{>}O(\frac{\sqrt{m}}{\sqrt{SK}}) \notag\\
r_8 & = \frac{(1-\lambda)^2(1+\lambda K -\lambda)m}{144K}\ge \frac{(1-\lambda)^2\lambda m}{144} = O(m) >O(\frac{\sqrt{m}}{\sqrt{SK}}) \notag,
\end{align}
where (a) follows from $ K \ge 1 + \lambda K -\lambda$ and (b) follows from $SK\ge m^{1/3}$.
Then we can set $\eta = O(\sqrt{m}/\sqrt{SK})$ and have the following convergence bound:
\begin{align}
&\frac{1}{SK} \sum_{s\!=\! 0}^{S\!-\!1}\sum_{k\!=\! 0}^{K\!-\!1}\Eb[\|\nabla f(\xb_{s,k})\|^2 \!+\!\frac{L^2}{m}\|\Q\x_{s,k}\|^2]  \notag\\
\!\stackrel{}{\le} &  O\Big(\frac{2\Eb[\mathfrak{P}_{0,0}\!-\!\mathfrak{P}_{S,0}]}{\sqrt{SKm}} + \frac{3L\sigma^2}{\sqrt{SKm}}  
 + \frac{72\sigma^2}{(1\!-\!\lambda)^2\sqrt{SKm}}\!+\! \frac{72\sigma^2}{(1\!-\!\lambda)^3\sqrt{SKm}} \Big). \notag
\end{align}
This completes the proof.
\end{proof}
  
  \onecolumn
  \clearpage
\section{Proof of main results}

First, we give the gloal view of our algorithm with matrix-vector formulation in Algorithm \ref{Algorithm: Global}.

For notational convenience, we define $\widetilde{\W} = \W \otimes \I_m,$ $\g_{s,k}^{(i)} = \nabla f_i(\x_{s,k}^{(i)};\vzeta_{s,k}^{(i)}),$, $\nf_{s,k}^{(i)} = \nabla f(\x_{s,k}^{(i)}),$  and $\a_{s,k} = [\a_{s,k}^{(i)\top},\cdots,\a_{s,k}^{(i)\top}]^\top$ and $\bar{\a}_{s,k}=\frac{1}{m} \sum_{i=1}^{m}\a_{s,k}^{(i)},$ for $\a \in \{\x,\y, \g,\nf\}$.
Here $\by_{s,k} = \bg_{s,k}$ because of $\by_{s,0} =  \bg_{s,0}.$ 

\begin{algorithm}[H]
\caption{Federated Network Learning with Gradient Tracking: Global View.}\label{Algorithm: Global}.
\begin{algorithmic} [1]
\STATE Set $\x_{0,0} = \x^0$ and $\y_{0,0} = \g_{0,0}$.
\FOR{$s = 0, \cdots, S-1$}
    \STATE  $\x_{s,1} \! =\! \widetilde{\W} \x_{s,0} - \eta \y_{s,0}$
	\STATE  $\y_{s,1} \!=\! \widetilde{\W}\y_{s,0}  + \g_{s,1}- \g_{s,0}$
	\FOR{$k = 1, \cdots, K-1$}
    	\STATE  $\x_{s,k+1} \! =\!  \x_{s,k} - \eta \y_{s,k}$
		\STATE  $\y_{s,k+1} \!=\!   \y_{s,k}  + \g_{s,k+1}- \g_{s,k}$
	\ENDFOR
	\STATE Set $\x_{s+1,0} = \x_{s,K}$ and $\y_{s+1,0} = \y_{s,K}$.
\ENDFOR
\end{algorithmic}
\end{algorithm}

\begin{customlemma}{1}[Descend Lemma]
Under Assumption, by applying Algorithm, we have the following inequality for any $s$:
\begin{align}
\Eb[f(\xb_{s,K}) -f(\xb_{s,0})]
\le 
&
- \frac{\eta}{2} \sum_{k= 0}^{K-1}\Eb[\|\nabla f(\xb_{s,k})\|^2] 
 - \frac{\eta}{2} \sum_{k= 0}^{K-1}\Eb[\|\bnf_{s,k}\|^2] \notag\\
 &
 + \frac{L\eta^2}{2} \sum_{k= 0}^{K-1}\Eb[\|\bg_{s,k}\|^2] + \frac{L^2\eta}{2m} \sum_{k= 0}^{K-1}\Eb[\|\Q\x_{s,k}\|^2 ]
\end{align}
\end{customlemma}

\begin{proof}
From the $L$-smoothness of $f$ and $\xb_{s,k+1}= \xb_{s,k} - \eta \by_{s,k} = \xb_{k} - \eta \bg_{s,k},$ we have
\begin{align}
f(\xb_{s,k+1}) &
\le
f(\xb_{s,k}) -  \langle \nabla f(\xb_{s,k}) , \xb_{s,k+1}- \xb_{s,k} \rangle + \frac{L}{2}\|\xb_{s,k+1}- \xb_{s,k}\|^2 \notag\\
&
=
f(\xb_{s,k}) - \eta \langle \nabla f(\xb_{s,k}) , \by_{s,k} \rangle + \frac{L\eta^2}{2}\|\by_{s,k}\|^2 \notag\\
&
=
f(\xb_{s,k}) - \eta \langle \nabla f(\xb_{s,k}) , \bg_{s,k} \rangle + \frac{L\eta^2}{2}\|\bg_{s,k}\|^2 
\end{align}
Due to $\Eb[\g_{s,k}^{(i)}| \Fc_{s,k}] = \nabla f_{s,k}^{(i)}$, we have 
\begin{align}
&
\Eb[f(\xb_{s,k+1})| \Fc_{s,k}] 
\le
f(\xb_{s,k}) - \eta \Eb[\langle \nabla f(\xb_{s,k}) , \bg_{s,k} \rangle| \Fc_{s,k}] + \frac{L\eta^2}{2}\Eb[\|\bg_{s,k}\|^2| \Fc_{s,k}] \notag\\
&
\le
f(\xb_{s,k}) - \eta \langle \nabla f(\xb_{s,k}) , \bnf_{s,k} \rangle + \frac{L\eta^2}{2}\Eb[\|\bg_{s,k}\|^2| \Fc_{s,k}] \notag\\
&
=
f(\xb_{s,k}) - \frac{\eta}{2}\|\nabla f(\xb_{s,k})\|^2 - \frac{\eta}{2} \|\bnf_{s,k}\|^2 + \frac{\eta}{2}\|\nabla f(\xb_{s,k})-\bnf_{s,k}\|^2 + \frac{L\eta^2}{2}\Eb[\|\bg_{s,k}\|^2| \Fc_{s,k}] \notag\\
&
\le
f(\xb_{s,k}) - \frac{\eta}{2}\|\nabla f(\xb_{s,k})\|^2 - \frac{\eta}{2} \|\bnf_{s,k}\|^2 + \frac{L^2\eta}{2m}\|\Q\x_{s,k}\|^2 + \frac{L\eta^2}{2}\Eb[\|\bg_{s,k}\|^2| \Fc_{s,k}] 
\end{align}

Taking the full expectation on the above inequality and telescoping from $k = 0$ to $K-1$ yields:
\begin{align}
\Eb[f(\xb_{s,K}) -f(\xb_{s,0})]
\le 
&
- \frac{\eta}{2} \sum_{k= 0}^{K-1}\Eb[\|\nabla f(\xb_{s,k})\|^2] 
 - \frac{\eta}{2} \sum_{k= 0}^{K-1}\Eb[\|\bnf_{s,k}\|^2] \notag\\
 &
 + \frac{L\eta^2}{2} \sum_{k= 0}^{K-1}\Eb[\|\bg_{s,k}\|^2] 
 + \frac{L^2\eta}{2m} \sum_{k= 0}^{K-1}\Eb[\|\Q\x_{s,k}\|^2 ]
\end{align}
\end{proof}

\begin{customlemma}{2}[Iterates Contraction]
Under Assumption, we have the following contraction property of the iterates from Algorithm:
\begin{align}
\|\Q\x_{s,k}\|^2 
\le
&
\lambda\|\Q\x_{s,0}\|^2 + \frac{\eta^2k}{1-\lambda} \sum_{t=0}^{k-1}\|\Q\y_{s,t}\|^2,\\ 
\|\Q\y_{s,k}\|^2 
\le
&
\lambda\|\Q\y_{s,0} \|^2 
+ \frac{6m\sigma^2}{1-\lambda}
+
\frac{24L^2}{1-\lambda}\|\Q\x_{s,0}\|^2 \notag\\
&+ \frac{12k\eta^2L^2}{1-\lambda} \sum_{t=0}^{k-1}\|\Q\y_{s,t}\|^2 + \frac{12mk\eta^2L^2}{1-\lambda}\sum_{t= 0}^{k-1}\|\by_{s,t}\|^2.
\end{align}
Furthermore, it holds that
\begin{align}
\sum_{k=0}^{K-1}\|\Q\x_{s,k}\|^2 
\le
&
(1+\lambda(K-1))\|\Q\x_{s,0}\|^2 + \frac{\eta^2K^2}{1-\lambda} \sum_{k=0}^{K-1}\|\Q\y_{s,k}\|^2, \label{Eq: sum_Qx}\\
\sum_{k=0}^{K-1}\|\Q\y_{s,k}\|^2 
\le
&
(1+\lambda(K-1))\|\Q\y_{s,0} \|^2 
+ \frac{6mK\sigma^2}{1-\lambda}
+
\frac{24KL^2}{1-\lambda}\|\Q\x_{s,0}\|^2 \notag\\
&
+ \frac{12\eta^2K^2L^2}{1-\lambda} \sum_{k=0}^{K-1}\|\Q\y_{s,k}\|^2 + \frac{12m\eta^2K^2L^2}{1-\lambda}\sum_{t= 0}^{k-1}\|\by_{s,t}\|^2. \label{Eq: sum_Qy}
\end{align}
\end{customlemma}

\begin{proof}
First, for vector $\x_t,$ we have the following contraction:
\begin{align}
\|\widetilde{\W}\x_{t} -\1\otimes\xb_{t} \|^2 = \|\widetilde{\W}(\x_{t} -\1\otimes\xb_{t}) \|^2 \le \lambda^2\|\x_{t} -\1\otimes\xb_{t}\|^2,
\end{align}
This is because $\x_{t} -\1\otimes\xb_{t}$ is orthogonal to $\1,$ which is the eigenvector corresponding to the largest eigenvalue of $\widetilde{\W},$ and $\lambda = \max\{|\lambda_2|,|\lambda_m|\}.$

Note that $\x_{s,k} = \widetilde{\W} \x_{s,0} - \eta \sum_{t=0}^{k-1} \y_{s,t}$ and $\xb_{s,k} = \xb_{s,0} - \eta \sum_{t=0}^{k-1} \by_{s,t}$.
Thus, we have 
\begin{align}\label{Eq: Qx}
&
\|\Q\x_{s,k}\|^2 = \|\x_{s,k}-\1\otimes\xb_{s,k}\|^2
=
\|\widetilde{\W} \x_{s,0} - \eta \sum_{t=0}^{k-1} \y_{s,t}-\1\otimes(\xb_{s,0} - \eta \sum_{t=0}^{k-1} \by_{s,t})\|^2 \notag\\
&
\le
(1+c_1)\|\widetilde{\W}\x_{s,0} -\1\otimes\xb_{s,0} \|^2 
+ 
(1+\frac{1}{c_1}) \eta^2\|\sum_{t=0}^{k-1}\y_{s,t}-\1\otimes \by_{s,t}\|^2 \notag\\
&
\le
\lambda\|\x_{s,0} -\1\otimes\xb_{s,0} \|^2 
+ 
\frac{\eta^2}{1-\lambda}\|\sum_{t=0}^{k-1}\y_{s,t}-\1\otimes \by_{s,t}\|^2 \notag\\
&
\le
\lambda\|\Q\x_{s,0}\|^2 + \frac{\eta^2k}{1-\lambda} \sum_{t=0}^{k-1}\|\Q\y_{s,t}\|^2.  
\end{align}

For $\y_{s,k}$, because $\y_{s,k}\! =\!   \y_{s,k-1}  \!+\! \g_{s,k}\!- \!\g_{s,k-1} \!=\! \widetilde{\W} \y_{s,0} \!+\! \g_{s,k}\!-\! \g_{s,0}$ and $\by_{s,k}\! =  \!\by_{s,0} \!+ \!\bg_{s,k}\!-\! \bg_{s,0}$, thus it holds that 
\begin{align}\label{Eq: Qy}
&
\|\Q\y_{s,k}\|^2 = \|\y_{s,k}-\1\otimes\by_{s,k}\|^2
=
\|\widetilde{\W} \y_{s,0} + \g_{s,k}- \g_{s,0} -\1\otimes(\by_{s,0} + \bg_{s,k}- \bg_{s,0} )\|^2 \notag\\
&
\le
(1+c_1)\|\widetilde{\W}\y_{s,0} -\1\otimes\by_{s,0} \|^2 
+ 
(1+\frac{1}{c_1}) \|\g_{s,k}- \g_{s,0} -\1\otimes( \bg_{s,k}- \bg_{s,0} )\|^2 \notag\\
&
\le
\lambda\|\y_{s,0} -\1\otimes\by_{s,0} \|^2 
+ 
\frac{1}{1-\lambda} \|(\I-\frac{1}{n}\1\1^\top)(\g_{s,k}- \g_{s,0})\|^2 \notag\\
&
\le
\lambda\|\Q\y_{s,0}  \|^2 
+ 
\frac{1}{1-\lambda} \|\g_{s,k}- \g_{s,0}\|^2 \notag\\
&
\le
\lambda\|\Q\y_{s,0}  \|^2 
+ 
\frac{3}{1-\lambda}(\|\g_{s,k} - \nf_{s,k}\|^2+\|\nf_{s,k}- \nf_{s,0}\|^2 + \|\g_{s,0} - \nf_{s,0}\|^2)
\notag\\
&
\le
\lambda\|\Q\y_{s,0}  \|^2 
+ 
\frac{3}{1-\lambda}(2m\sigma^2+L^2\|\x_{s,k}- \x_{s,0}\|^2)
\end{align}

Note that for the term $\|\x_{s,k}- \x_{s,0}\|^2 $, it can be bounded as
\begin{align}\label{Eq: x-x}
&
\|\x_{s,k}- \x_{s,0}\|^2 
= 
\|\widetilde{\W} \x_{s,0}-  \eta \sum_{t=0}^{k-1} \y_{s,t} - \x_{s,0}\|^2 \notag\\
=
&
\|(\widetilde{\W} -\I)\x_{s,0} - \eta \sum_{t=0}^{k-1} \y_{s,t}\|^2
\le 2\|(\widetilde{\W} -\I)\x_{s,0} \|^2 + 2\eta^2 \|\sum_{t=0}^{k-1} \y_{s,t}\|^2 \notag\\
=
&
2\|(\widetilde{\W} -\I)(\x_{s,0} - \1\otimes\xb_{s,0}) \|^2 + 2\eta^2 \|\sum_{t=0}^{k-1} \y_{s,t}\|^2 \notag\\
\le
&
2\|(\widetilde{\W} -\I)(\x_{s,0} - \1\otimes\xb_{s,0}) \|^2 + 2\eta^2k\sum_{t=0}^{k-1} \| \y_{s,t}\|^2 \notag\\
\stackrel{}{\le}
&
8\| \x_{s,0} - \1\otimes\xb_{s,0} \|^2 + 2\eta^2k\sum_{t=0}^{k-1} \| \y_{s,t}\|^2 \notag\\
\stackrel{(a)}{\le}
&
8\| \x_{s,0} - \1\otimes\xb_{s,0} \|^2 + 2\eta^2k\sum_{t=0}^{k-1} (2\| \y_{s,t} - \1\otimes \by_{s,t}\|^2 +2 \|\1\otimes \by_{s,t}\|^2) \notag\\
\stackrel{}{\le}
&
8\|\Q\x_{s,0} \|^2 + 4\eta^2k\sum_{t=0}^{k-1} \| \Q\y_{s,t}\|^2  + 4\eta^2mk\sum_{t=0}^{k-1} \| \by_{s,t}\|^2
\end{align}
where (a) is due to $\|\widetilde{\W} -\I\| \le 2.$

Thus, by plugging~\eqref{Eq: x-x} into~\eqref{Eq: Qy}, we have  
\begin{align}
\|\Q\y_{s,k}\|^2 
\le
&\lambda\|\Q\y_{s,0} \|^2 
+ \frac{6m\sigma^2}{1-\lambda}
+
\frac{24L^2}{1-\lambda}\|\Q\x_{s,0}\|^2 \notag\\
&+ \frac{12k\eta^2L^2}{1-\lambda} \sum_{t=0}^{k-1}\|\Q\y_{s,t}\|^2 + \frac{12mk\eta^2L^2}{1-\lambda}\sum_{t= 0}^{k-1}\|\by_{s,t}\|^2
\end{align}

Furthermore, by telescoping~\eqref{Eq: Qx} from $k=0$ to $K-1$, we have 
\begin{align}
\sum_{k=0}^{K-1}\|\Q\x_{s,k}\|^2 
&\le
(1+\lambda(K-1))\|\Q\x_{s,0}\|^2 + \frac{\eta^2K}{1-\lambda} \sum_{k=0}^{K-1}\sum_{t=0}^{k-1}\|\Q\y_{s,t}\|^2 \notag\\
&
\le
(1+\lambda(K-1))\|\Q\x_{s,0}\|^2 + \frac{\eta^2K^2}{1-\lambda} \sum_{k=0}^{K-1}\|\Q\y_{s,k}\|^2.
\end{align}
Similarly, it holds that
\begin{align}
\sum_{k=0}^{K-1}\|\Q\y_{s,k}\|^2 
\le
&
(1+\lambda(K-1))\|\Q\y_{s,0} \|^2 
+ \frac{6mK\sigma^2}{1-\lambda}
+
\frac{24KL^2}{1-\lambda}\|\Q\x_{s,0}\|^2 \notag\\
&
+ \frac{12\eta^2L^2}{1-\lambda} \sum_{k=0}^{K-1}k\sum_{t=0}^{k-1}\|\Q\y_{s,t}\|^2 + \frac{12m\eta^2L^2}{1-\lambda}\sum_{k=0}^{K-1}k\sum_{t= 0}^{k-1}\|\by_{s,t}\|^2\notag\\
\le
&
(1+\lambda(K-1))\|\Q\y_{s,0} \|^2 
+ \frac{6mK\sigma^2}{1-\lambda}
+
\frac{24KL^2}{1-\lambda}\|\Q\x_{s,0}\|^2 \notag\\
&
+ \frac{12\eta^2K^2L^2}{1-\lambda} \sum_{k=0}^{K-1}\|\Q\y_{s,k}\|^2 + \frac{12m\eta^2K^2L^2}{1-\lambda}\sum_{t= 0}^{k-1}\|\by_{s,t}\|^2
\end{align}
\end{proof}

\subsection{Proof Details for Theorem~\ref{thm:Convergence of NET-FLEET}}
\begin{proof}
From Lemma~\ref{Lem: Descend}, we have 
\begin{align}
\Eb[f(\xb_{s,K}) -f(\xb_{s,0})]
\le 
&
- \frac{\eta}{2} \sum_{k= 0}^{K-1}\Eb[\|\nabla f(\xb_{s,k})\|^2] 
 - \frac{\eta}{2} \sum_{k= 0}^{K-1}\Eb[\|\bnf_{s,k}\|^2] - \frac{L^2\eta}{2m} \sum_{k= 0}^{K-1}\Eb[\|\Q\x_{s,k}\|^2 ] \notag\\
 &
 + \frac{L\eta^2}{2} \sum_{k= 0}^{K-1}\Eb[\|\bg_{s,k}\|^2] + \frac{L^2\eta}{m} \sum_{k= 0}^{K-1}\Eb[\|\Q\x_{s,k}\|^2 ].
\end{align}
Plugging~\eqref{Eq: sum_Qx} from Lemma~\ref{Lem: Iterates}, we have 
\begin{align}\label{Eq: diff_f}
&\Eb[f(\xb_{s,K}) \!-\!f(\xb_{s,0})]
\!\le\! 
- \frac{\eta}{2} \sum_{k= 0}^{K-1}\Eb[\|\nabla f(\xb_{s,k})\|^2] 
 - \frac{\eta}{2} \sum_{k= 0}^{K-1}\Eb[\|\bnf_{s,k}\|^2]- \frac{L^2\eta}{2m} \sum_{k= 0}^{K-1}\Eb[\|\Q\x_{s,k}\|^2 ]\notag\\
 &
 + \frac{L\eta^2}{2} \sum_{k= 0}^{K-1}\Eb[\|\bg_{s,k}\|^2]  
 + \frac{L^2\eta}{m}(1+\lambda(K-1))\Eb\|\Q\x_{s,0}\|^2 + \frac{\eta^3L^2K^2}{m(1-\lambda)} \sum_{k=0}^{K-1}\Eb\|\Q\y_{s,k}\|^2.
\end{align}

Also, from Lemma~\ref{Lem: Iterates}, for some constant $C_1$, it holds that 
\begin{align}\label{Eq: diff_p}
&(\|\Q\x_{s,K}\|^2 + C_1\eta^2\|\Q\x_{s,K}\|^2) - 
(\|\Q\x_{s,0}\|^2 + C_1\eta^2\|\Q\x_{s,0}\|^2) \notag\\
\le
&
-(1-\lambda - \frac{24C_1L^2\eta^2}{1-\lambda})\|\Q\x_{s,0}\|^2
-(1-\lambda)C_1\eta^2\|\Q\y_{s,0}\|^2 \notag\\
&
+\! \frac{\eta^2K \!+ \!12 C_1KL^2\eta^4}{1\!-\!\lambda} \sum_{k=0}^{K-1}\|\Q\y_{s,k}\|^2 
\!+\! \frac{12mC_1L^2K^2\eta^4}{1\!-\!\lambda}\sum_{k=0}^{K-1}\|\by_{s,k}\|^2 
\!+\! \frac{6mC_1\eta^2\sigma^2}{1\!-\!\lambda}
\end{align}

Thus, combining~\eqref{Eq: diff_f} and~\eqref{Eq: diff_p}, we have
\begin{align}\label{Eq: Desend_fp}
&\Eb[f(\xb_{s,K}) \!-\!f(\xb_{s,0}) \!+\! \frac{1}{mK}(\|\Q\x_{s,K}\|^2 \!+\! C_1\eta^2\|\Q\x_{s,K}\|^2) \!-\! 
\frac{1}{mK}(\|\Q\x_{s,0}\|^2 \!+\! C_1\eta^2\|\Q\x_{s,0}\|^2)]\notag\\
\!\le\! 
&
- \frac{\eta}{2} \sum_{k= 0}^{K-1}\Eb[\|\nabla f(\xb_{s,k})\|^2] 
 - \frac{\eta}{2} \sum_{k= 0}^{K-1}\Eb[\|\bnf_{s,k}\|^2]- \frac{L^2\eta}{2m} \sum_{k= 0}^{K-1}\Eb[\|\Q\x_{s,k}\|^2 ]\notag\\
 &
 + \frac{L\eta^2}{2} \sum_{k= 0}^{K-1}\Eb[\|\bg_{s,k}\|^2]  
 + \frac{L^2\eta}{m}(1+\lambda(K-1))\Eb\|\Q\x_{s,0}\|^2 + \frac{\eta^3L^2K^2}{m(1-\lambda)} \sum_{k=0}^{K-1}\Eb\|\Q\y_{s,k}\|^2\notag\\
 &
 -(1-\lambda - \frac{24C_1L^2\eta^2}{1-\lambda})\frac{1}{mK}\|\Q\x_{s,0}\|^2
-(1-\lambda)\frac{C_1\eta^2}{mK}\|\Q\y_{s,0}\|^2 \notag\\
&
+\! \frac{\eta^2 \!+ \!12 C_1L^2\eta^4}{(1\!-\!\lambda)m} \sum_{k=0}^{K-1}\|\Q\y_{s,k}\|^2 
\!+\! \frac{12C_1L^2K\eta^4}{(1\!-\!\lambda)}\sum_{k=0}^{K-1}\|\by_{s,k}\|^2 
\!+\! \frac{6C_1\eta^2\sigma^2}{(1\!-\!\lambda)K}\notag\\
\!=\! 
&
- \frac{\eta}{2} \sum_{k= 0}^{K-1}\Eb[\|\nabla f(\xb_{s,k})\|^2] 
 - \frac{\eta}{2} \sum_{k= 0}^{K-1}\Eb[\|\bnf_{s,k}\|^2]- \frac{L^2\eta}{2m} \sum_{k= 0}^{K-1}\Eb[\|\Q\x_{s,k}\|^2 ]
\!+\! \frac{6C_1\eta^2\sigma^2}{(1\!-\!\lambda)K}\notag\\
 &
 + (\frac{L\eta^2}{2}+\frac{12C_1L^2K\eta^4}{(1\!-\!\lambda)}) \sum_{k= 0}^{K-1}\Eb[\|\bg_{s,k}\|^2]  
 + (\frac{\eta^3L^2K^2}{m(1-\lambda)} + \frac{\eta^2 \!+ \!12 C_1L^2\eta^4}{(1\!-\!\lambda)m}) \sum_{k=0}^{K-1}\Eb\|\Q\y_{s,k}\|^2\notag\\
 &
 -(1-\lambda - \frac{24C_1L^2\eta^2}{1-\lambda} - (1+\lambda(K-1))KL^2\eta)\frac{1}{mK}\Eb[\|\Q\x_{s,0}\|^2]-(1-\lambda)\frac{C_1\eta^2}{mK}\Eb[\|\Q\y_{s,0}\|^2] \notag\\
\!\le\! 
&
- \frac{\eta}{2} \sum_{k= 0}^{K-1}\Eb[\|\nabla f(\xb_{s,k})\|^2] 
 - \frac{\eta}{2} \sum_{k= 0}^{K-1}\Eb[\|\bnf_{s,k}\|^2]- \frac{L^2\eta}{2m} \sum_{k= 0}^{K-1}\Eb[\|\Q\x_{s,k}\|^2 ]
\!+\! \frac{6C_1\eta^2\sigma^2}{(1\!-\!\lambda)K}\notag\\
 &
 + (\frac{L\eta^2}{2}+\frac{12C_1L^2K\eta^4}{(1\!-\!\lambda)}) \sum_{k= 0}^{K-1}\Eb[\|\bg_{s,k}\|^2]  
 + \frac{3\eta^2}{(1\!-\!\lambda)m} \sum_{k=0}^{K-1}\Eb\|\Q\y_{s,k}\|^2\notag\\
 &
 -(1-\lambda - \frac{24C_1L^2\eta^2}{1-\lambda} - (1+\lambda(K-1))KL^2\eta)\frac{1}{mK}\Eb[\|\Q\x_{s,0}\|^2]-(1-\lambda)\frac{C_1\eta^2}{mK}\Eb[\|\Q\y_{s,0}\|^2] 
\end{align}

From\eqref{Eq: sum_Qy} in Lemma~\ref{Lem: Iterates}, we have 
\begin{align}
(1- \frac{12\eta^2K^2L^2}{1-\lambda})\sum_{k=0}^{K-1}\|\Q\y_{s,k}\|^2 
\le
&
(1+\lambda(K-1))\|\Q\y_{s,0} \|^2 
+ \frac{6mK\sigma^2}{1-\lambda}\notag\\
&
+
\frac{24KL^2}{1-\lambda}\|\Q\x_{s,0}\|^2 
+ \frac{12m\eta^2K^2L^2}{1-\lambda}\sum_{t= 0}^{k-1}\|\by_{s,t}\|^2,
\end{align}
which implies that 
\begin{align}\label{Eq: sum_Qy2}
\sum_{k=0}^{K-1}\|\Q\y_{s,k}\|^2 
\le
&
2(1+\lambda(K-1))\|\Q\y_{s,0} \|^2 
+ \frac{12mK\sigma^2}{1-\lambda}\notag\\
&
+
\frac{48KL^2}{1-\lambda}\|\Q\x_{s,0}\|^2 
+ \frac{24m\eta^2K^2L^2}{1-\lambda}\sum_{t= 0}^{k-1}\|\by_{s,t}\|^2.
\end{align}

By plugging~\eqref{Eq: sum_Qy2} into~\eqref{Eq: Desend_fp}, we have 
\begin{align}
&\Eb[f(\xb_{s,K}) \!-\!f(\xb_{s,0}) + \frac{1}{mK}(\|\Q\x_{s,K}\|^2 + C_1\eta^2\|\Q\x_{s,K}\|^2) - 
\frac{1}{mK}(\|\Q\x_{s,0}\|^2 + C_1\eta^2\|\Q\x_{s,0}\|^2)]\notag\\
\!\le\!
&
- \frac{\eta}{2} \sum_{k= 0}^{K-1}\Eb[\|\nabla f(\xb_{s,k})\|^2] 
 - \frac{\eta}{2} \sum_{k= 0}^{K-1}\Eb[\|\bnf_{s,k}\|^2]- \frac{L^2\eta}{2m} \sum_{k= 0}^{K-1}\Eb[\|\Q\x_{s,k}\|^2 ]
\!+\! \frac{6C_1\eta^2\sigma^2}{(1\!-\!\lambda)K}\notag\\
 &
 + (\frac{L\eta^2}{2}+\frac{12C_1L^2K\eta^4}{(1\!-\!\lambda)}) \sum_{k= 0}^{K-1}\Eb[\|\bg_{s,k}\|^2]  
 + \frac{3\eta^2}{(1\!-\!\lambda)m}\Big(2(1+\lambda(K-1))\|\Q\y_{s,0} \|^2 \notag\\
&
+ \frac{12mK\sigma^2}{1-\lambda}
+
\frac{48KL^2}{1-\lambda}\|\Q\x_{s,0}\|^2 
+ \frac{24m\eta^2K^2L^2}{1-\lambda}\sum_{t= 0}^{k-1}\|\by_{s,t}\|^2\Big)\notag\\
 &
 -(1-\lambda - \frac{24C_1L^2\eta^2}{1-\lambda} - (1+\lambda(K-1))KL^2\eta)\frac{1}{mK}\Eb[\|\Q\x_{s,0}\|^2]-(1-\lambda)\frac{C_1\eta^2}{mK}\Eb[\|\Q\y_{s,0}\|^2] \notag\\
\!=\!
&
- \frac{\eta}{2} \sum_{k= 0}^{K-1}\Eb[\|\nabla f(\xb_{s,k})\|^2] 
 - \frac{\eta}{2} \sum_{k= 0}^{K-1}\Eb[\|\bnf_{s,k}\|^2]- \frac{L^2\eta}{2m} \sum_{k= 0}^{K-1}\Eb[\|\Q\x_{s,k}\|^2 ]
\notag\\
 &
 + (\frac{L\eta^2}{2}+\frac{12C_1L^2K\eta^4}{(1\!-\!\lambda)} + \frac{72L^2K^2\eta^4}{(1-\lambda)^2}) \sum_{k= 0}^{K-1}\Eb[\|\bg_{s,k}\|^2]  
 + \frac{36K\eta^2\sigma^2}{(1\!-\!\lambda)^2}\!+\! \frac{6C_1\eta^2\sigma^2}{(1\!-\!\lambda)K}\notag\\
&
 -(1-\lambda - \frac{24C_1L^2\eta^2}{1-\lambda} - (1+\lambda(K-1))KL^2\eta - \frac{144L^2\eta^2}{(1-\lambda)^2})\frac{1}{mK}\Eb[\|\Q\x_{s,0}\|^2] \notag\\
 &
 -(1-\lambda - \frac{6(1+\lambda K -\lambda)}{C_1(1-\lambda)})\frac{C_1\eta^2}{mK}\Eb[\|\Q\y_{s,0}\|^2]  \notag\\
\!\le\!
&
- \frac{\eta}{2} \sum_{k= 0}^{K-1}\Eb[\|\nabla f(\xb_{s,k})\|^2] 
- \frac{L^2\eta}{2m} \sum_{k= 0}^{K-1}\Eb[\|\Q\x_{s,k}\|^2 ]
\notag\\
 &
  - \Big( 1- L\eta-\frac{24C_1L^2K\eta^3}{(1\!-\!\lambda)} - \frac{144L^2K^2\eta^3}{(1-\lambda)^2} \Big) \frac{\eta}{2}\sum_{k= 0}^{K-1}\Eb[\|\bnf_{s,k}\|^2]\notag\\
&
 -(1-\lambda - \frac{24C_1L^2\eta^2}{1-\lambda} - (1+\lambda(K-1))KL^2\eta - \frac{144L^2\eta^2}{(1-\lambda)^2})\frac{1}{mK}\Eb[\|\Q\x_{s,0}\|^2] \notag\\
 &
 -(1-\lambda - \frac{6(1+\lambda K -\lambda)}{C_1(1-\lambda)})\frac{C_1\eta^2}{mK}\Eb[\|\Q\y_{s,0}\|^2]  \notag\\
 &
 + (\frac{L\eta^2}{2}+\frac{12C_1L^2K\eta^4}{(1\!-\!\lambda)} + \frac{72L^2K^2\eta^4}{(1-\lambda)^2}) \frac{K\sigma^2}{m}  
 + \frac{36K\eta^2\sigma^2}{(1\!-\!\lambda)^2}\!+\! \frac{6C_1\eta^2\sigma^2}{(1\!-\!\lambda)K}
\end{align}

Define the constants 
\begin{align}
&C_{\nf} \triangleq 1- L\eta-\frac{24C_1L^2K\eta^3}{(1\!-\!\lambda)} - \frac{144L^2K^2\eta^3}{(1-\lambda)^2} ,\\
&C_{\x} \triangleq 1-\lambda - \frac{24C_1L^2\eta^2}{1-\lambda} - (1+\lambda(K-1))KL^2\eta - \frac{144L^2\eta^2}{(1-\lambda)^2}, \\
&C_{\y} \triangleq 1-\lambda - \frac{6(1+\lambda K -\lambda)}{C_1(1-\lambda)}.
\end{align}
By setting $C_1 = 6(1+\lambda K-\lambda) / (1-\lambda)^2$, we have $C_{\y} = 0$. Then, the other constants are
\begin{align}
&C_{\nf} = 1- L\eta-\frac{144(1+\lambda K-\lambda)L^2K\eta^3}{(1\!-\!\lambda)^3} - \frac{144L^2K^2\eta^3}{(1-\lambda)^2}, \\
&C_{\x} = 1-\lambda - \frac{144(1+\lambda K-\lambda)L^2\eta^2}{(1-\lambda)^3} - (1+\lambda(K-1))KL^2\eta - \frac{144L^2\eta^2}{(1-\lambda)^2}. 
\end{align}
By letting $\eta \le \min\{\sqrt{(1-\lambda)^3/144(1+\lambda K -\lambda)LK}, \sqrt{(1-\lambda)^2/144LK^2}, 1/3L\}$, we have $C_{\nf} \ge 1- 3 L\eta \ge 0$.
Also, letting $\eta \le \min\{(1-\lambda)/3(1+\lambda K -\lambda)L^2K, (1-\lambda)^3K/144, (1-\lambda)^2(1+\lambda K -\lambda)K/144\}$, we have $C_{\x} \ge 3(1+\lambda(K-1))KL^2\eta \ge 0$.

With the above parameter setting and  the potential function $\mathfrak{P}_{s,k} \!\triangleq\! f(\xb_{s,k}) + \frac{1}{mK}(\|\Q\x_{s,k}\|^2 + C_1\eta^2\|\Q\x_{s,k}\|^2)$, we have 
\begin{align}
&\frac{\eta}{2} \sum_{k= 0}^{K-1}\Eb[\|\nabla f(\xb_{s,k})\|^2 +\frac{L^2}{m}\|\Q\x_{s,k}\|^2]  \notag\\
\le&  \Eb[\mathfrak{P}_{s,0}-\mathfrak{P}_{s,K}] + (\frac{L\eta^2}{2}+\frac{12C_1L^2K\eta^4}{(1\!-\!\lambda)} + \frac{72L^2K^2\eta^4}{(1-\lambda)^2}) \frac{K\sigma^2}{m}  
 + \frac{36K\eta^2\sigma^2}{(1\!-\!\lambda)^2}\!+\! \frac{6C_1\eta^2\sigma^2}{(1\!-\!\lambda)K} 
\end{align}

Telescope $s$ from $0$ to $S-1$, we have 
\begin{align}
&\frac{\eta}{2} \sum_{s= 0}^{S-1}\sum_{k= 0}^{K-1}\Eb[\|\nabla f(\xb_{s,k})\|^2 +\frac{L^2}{m}\|\Q\x_{s,k}\|^2]  \notag\\
\le&  \Eb[\mathfrak{P}_{0,0}-\mathfrak{P}_{0,K}] + \Eb[\mathfrak{P}_{1,0}-\mathfrak{P}_{1,K}] + \cdots + \Eb[\mathfrak{P}_{S-1,0}-\mathfrak{P}_{S-1,K}] \notag\\
&+ (\frac{L\eta^2}{2}+\frac{12C_1L^2K\eta^4}{(1\!-\!\lambda)} + \frac{72L^2K^2\eta^4}{(1-\lambda)^2}) \frac{SK\sigma^2}{m}  
 + \frac{36SK\eta^2\sigma^2}{(1\!-\!\lambda)^2}\!+\! \frac{6C_1\eta^2\sigma^2S}{(1\!-\!\lambda)K} \notag\\
= &  \Eb[\mathfrak{P}_{0,0}\!-\!\mathfrak{P}_{S,0}] \!+\! (\frac{L\eta^2}{2}+\frac{12C_1L^2K\eta^4}{(1\!-\!\lambda)} \!+\! \frac{72L^2K^2\eta^4}{(1-\lambda)^2}) \frac{SK\sigma^2}{m}  
 \!+\! \frac{36SK\eta^2\sigma^2}{(1\!-\!\lambda)^2}\!+\! \frac{6C_1\eta^2\sigma^2S}{(1\!-\!\lambda)K}
\end{align}

Multiplying the factor $2SK/\eta$ at both sides, we have 
\begin{align}
&\frac{1}{SK} \sum_{s\!=\! 0}^{S\!-\!1}\sum_{k\!=\! 0}^{K\!-\!1}\Eb[\|\nabla f(\xb_{s,k})\|^2 \!+\!\frac{L^2}{m}\|\Q\x_{s,k}\|^2]  \notag\\
\!\le\! &  \frac{2\Eb[\mathfrak{P}_{0,0}\!-\!\mathfrak{P}_{S,0}]}{SK\eta} \!+\! (\frac{L\eta}{2}\!+\!\frac{72(1\!+\!\lambda K \!-\!\lambda)L^2K\eta^3}{(1\!-\!\lambda)^3} \!+\! \frac{72L^2K^2\eta^3}{(1\!-\!\lambda)^2}) \frac{2\sigma^2}{m}  
 \!+\! \frac{72\eta\sigma^2}{(1\!-\!\lambda)^2}\!+\! \frac{72(1\!+\!\lambda K \!-\!\lambda)\eta\sigma^2}{(1\!-\!\lambda)^3K^2} \notag\\
\stackrel{(a)}{\le} &  \frac{2\Eb[\mathfrak{P}_{0,0}\!-\!\mathfrak{P}_{S,0}]}{SK\eta} + (\frac{L\eta}{2}\!+\!\frac{72\eta^2}{(1\!-\!\lambda)^2} \!+\! \frac{72L^2K^2\eta^3}{(1\!-\!\lambda)^2}) \frac{2\sigma^2}{m}  
 \!+\! \frac{72\eta\sigma^2}{(1\!-\!\lambda)^2}\!+\! \frac{72\eta\sigma^2}{(1\!-\!\lambda)^3K}
\end{align}
where (a) is by $\eta \le (1-\lambda)/3(1+\lambda K -\lambda) KL^2$ and $1 + K\lambda - \lambda \le K$.
\end{proof}

\subsection{Proof Details of Corollary~\ref{cor: linear speedup}}
\begin{proof}
Setting $\eta = O(\sqrt{SK})$ and $K = \sqrt[4]{SK}$ (i.e. $K = S^{1/3}$), we have 
\begin{align}
&\frac{1}{SK} \sum_{s\!=\! 0}^{S\!-\!1}\sum_{k\!=\! 0}^{K\!-\!1}\Eb[\|\nabla f(\xb_{s,k})\|^2 \!+\!\frac{L^2}{m}\|\Q\x_{s,k}\|^2]  \notag\\
\!\stackrel{}{\le}\! &  O\Big(\frac{2\Eb[\mathfrak{P}_{0,0}\!-\!\mathfrak{P}_{S,0}]}{\sqrt{SK}} + (\frac{L}{2\sqrt{SK}}\!+\!\frac{72}{(1\!-\!\lambda)^2SK} \!+\! \frac{72L^2}{(1\!-\!\lambda)^2SK}) \frac{2\sigma^2}{m}  
 \!+\! \frac{72\sigma^2}{(1\!-\!\lambda)^2\sqrt{SK}}\!+\! \frac{72\sigma^2}{(1\!-\!\lambda)^3(SK)^{3/4}}\Big) \notag\\
 =& O\Big(\frac{2\Eb[\mathfrak{P}_{0,0}\!-\!\mathfrak{P}_{S,0}]}{\sqrt{SK}} + (\frac{1}{\sqrt{SK}}\!+\!\frac{1}{(1\!-\!\lambda)^2SK} ) \frac{\sigma^2}{m}  
 \!+\! \frac{\sigma^2}{(1\!-\!\lambda)^2\sqrt{SK}}\!+\! \frac{\sigma^2}{(1\!-\!\lambda)^3(SK)^{3/4}}\Big)
\end{align}
If we ignore the factor caused by the network topolgy and worker number, then the convergence rate is 
\begin{align}
O\Big(\frac{2\Eb[\mathfrak{P}_{0,0}\!-\!\mathfrak{P}_{S,0}]}{\sqrt{SK}}  
 \!+\! \frac{\sigma^2}{\sqrt{SK}}\Big),
\end{align}
which matches the results for vanilla SGD.

Furthermore, we check the above paramter settings are valid.
Recall that the condition on $\eta$ is
\begin{align}
\eta \le \min\{&\underbrace{\sqrt{\frac{(1-\lambda)^3}{144(1+\lambda K -\lambda)LK}}}_{r_1}, \underbrace{\sqrt{\frac{(1-\lambda)^2}{144LK^2}}}_{r_2}, \frac{1}{3L}, \notag\\
&\underbrace{\frac{(1-\lambda)}{3(1+\lambda K -\lambda)L^2K}}_{r_3}, \underbrace{\frac{(1-\lambda)^3K}{144}}_{r_4}, \underbrace{\frac{(1-\lambda)^2(1+\lambda K -\lambda)K}{144}}_{r_5}\}.
\end{align}
Plugging $K= \sqrt[4]{SK}$, we have $r_1$ and $r_2$ are the order of $O(\frac{1}{K}) = O(\frac{1}{\sqrt[4]{SK}})$, which is larger than $O(\frac{1}{\sqrt{SK}})$; $r_3$ is the order of $O(\frac{1}{K^2}) = O(\frac{1}{\sqrt{SK}})$; $r_4$ and $r_5$ are the order of $\Omega(1)$. Thus, $\eta = O(\frac{1}{\sqrt{SK}})$ and $K = \sqrt[4]{SK}$ are valid.

\end{proof}

\end{document}